\documentclass{article}
\PassOptionsToPackage{numbers,compress}{natbib}
\PassOptionsToPackage{table,dvipsnames}{xcolor}
\usepackage[preprint]{neurips_2026}
\usepackage[utf8]{inputenc} 
\usepackage[T1]{fontenc}    
\usepackage{hyperref}       
\usepackage{url}            
\usepackage{booktabs}       
\usepackage{amsfonts,amssymb}       
\usepackage{nicefrac}       
\usepackage{microtype}      
\usepackage{graphicx}
\usepackage{amsmath}
\usepackage{adjustbox}
\usepackage{multirow}
\usepackage{bm}
\usepackage{subcaption}
\usepackage{array}
\usepackage{makecell}
\usepackage{wrapfig}
\usepackage{longtable}
\usepackage{seqsplit}
\usepackage{enumitem}
\usepackage{array}
\usepackage{ragged2e}
\usepackage{tikz}
\usepackage{makecell}
\usetikzlibrary{arrows.meta}

\newcommand{\ie}{\emph{i.e.}}    
\newcommand{\eg}{\emph{e.g.}}    

\DeclareMathOperator*{\argmax}{arg\,max}

\title{PrISM-IQA: Image Quality Assessment Made Practical for Smartphone Photography}

\author{
\textbf{Shuyan Zhai$^{1}$,
Jiaqi He$^{1}$,
Weixia Zhang$^{2}$,
Liang Wang$^{3}$,}\\
\textbf{Zhenjie Lee$^{3}$,
Zufeng Zhang$^{4}$,
Kede Ma$^{1}$\thanks{Corresponding author.}}\\[0.5em]
$^{1}$Department of Computer Science, City University of Hong Kong\\
$^{2}$School of Computer Science, Institute of AI, Shanghai Jiao Tong University\\
$^{3}$OPPO Research Institute\\
$^{4}$School of Vehicle and Mobility, Tsinghua University\\[0.5em]
\texttt{\{shuyazhai2-c,jiaqi.he\}@my.cityu.edu.hk,zwx8981@sjtu.edu.cn,}\\
\texttt{\{leon.wang,lizhenjie\}@oppo.com}\\
\texttt{zhangzufeng@tsari.tsinghua.edu.cn, kede.ma@cityu.edu.hk} \\
\url{https://github.com/Multimedia-Analytics-Laboratory/PrISM-IQA}
}

\begin{document}
\maketitle
\begin{abstract}
 Existing smartphone image quality assessment (IQA) methods commonly reduce perceptual quality to a single score. However, this scalar formulation is poorly aligned with practical image signal processor (ISP) tuning, where engineers must identify specific quality issues, estimate their severities, and determine whether they are acceptable or require intervention. In this work, we introduce a Practical ISP-aware Structured Model for IQA (PrISM-IQA), which reformulates smartphone IQA as a multi-issue ordinal diagnosis problem. Rather than regressing a single quality score, PrISM-IQA predicts an \textit{ordered} severity level---absent, minor, severe, or critical---for each ISP-relevant issue, covering both global image-level artifacts and local content-dependent defects. To produce logically consistent predictions, PrISM-IQA combines cumulative ordinal encoding with structured inference that captures within-issue monotonicity as well as cross-issue subsumption and exclusion relations. We evaluate PrISM-IQA on a reconstructed SPAQ benchmark annotated with $53$ ISP-relevant quality issues and on a small-scale expert-annotated real-world dataset. Experimental results demonstrate the effectiveness of PrISM-IQA for practical issue-level diagnosis, reveal transferable perceptual quality representations through linear probing, and further show how its predictions can support actionable and meaningful ISP tuning.
\end{abstract}

\section{Introduction}
Smartphone photography has become the dominant mode of image capture, creating growing demand for camera pipelines that can deliver perceptually faithful and pleasing images across diverse scenes and operating conditions~\cite{delbracio2021mobile}. In practice, much of this perceptual quality is shaped by the image signal processor (ISP), whose tuning governs exposure, color,  denoising, sharpening, tone mapping, and related transformations. Despite substantial progress in image quality assessment (IQA), however, the objectives of mainstream IQA research~\cite{wang2006modern} remain only weakly aligned with the needs of ISP development. Most existing no-reference IQA methods~\cite{ma2017end,mittal2012no,ye2012unsupervised,mittal2013niqe,su2020blindly,ke2021musiq,zhang2021uncertainty,zhang2023blind} are designed to predict a single quality score, typically in the form of a mean opinion score (MOS)-like scalar. By contrast, practical ISP tuning requires a more diagnostic output: which quality issue is present, how severe it is, and whether it is acceptable or requires corrective action.

This mismatch is not merely a matter of evaluation preference, but a limitation of the scalar formulation itself. A single quality score is useful for ranking images or approximating average human preference, yet it often obscures the useful information needed for engineering decisions. Two images with similar MOSs may arise from very different failure modes, such as under-exposure, color cast, loss of detail, or over-sharpening, each of which calls for a different tuning strategy. Moreover, ISP decisions are often made around operational thresholds rather than along a smooth perceptual continuum: an issue may be absent, tolerable, noticeable, or severe enough to trigger intervention. For smartphone photography, therefore, a practical IQA system should move beyond holistic scoring and instead provide issue-level diagnosis in a form that is directly useful for ISP optimization.

Motivated by this need, we reformulate smartphone IQA as a \textit{multi-issue ordinal diagnosis} problem and introduce  a \textbf{Pr}actical \textbf{I}SP-aware \textbf{S}tructured \textbf{M}odel for IQA (PrISM-IQA). Instead of regressing a single quality score, PrISM-IQA predicts an ordered severity level for each ISP-relevant issue, using four levels: \textit{absent}, \textit{minor}, \textit{severe}, and \textit{critical}. The issue space includes both global image-level problems and local content-dependent problems associated with specific semantic regions. This formulation yields outputs that are more interpretable than scalar quality prediction and better aligned with practical ISP workflows, where the central question is often not which image is slightly better overall, but whether a specific defect is present and actionable.

A further challenge is that the issue labels are not independent. Some labels exhibit subsumption relations, in which a more specific defect implies a broader parent issue. For example, over-saturation can be regarded as a specific instance of a broader color-related problem. Other labels may be mutually exclusive and therefore should not be predicted simultaneously; for instance, an image should not be diagnosed as both over-bright and under-bright at the same time. Ignoring this structure can produce predictions that are numerically plausible yet logically inconsistent. PrISM-IQA addresses this challenge by leveraging a \textit{structured prediction} layer~\cite{deng2014large}, tailored to smartphone photography quality diagnosis. By combining cumulative ordinal encoding~\cite{li2006ordinal} with structured inference, PrISM-IQA respects both the ordered nature of severity labels and the semantic dependencies among issues, yielding logically consistent and practically meaningful diagnosis.

We evaluate PrISM-IQA in two complementary settings. First, we reconstruct the public SPAQ dataset~\cite{fang2020perceptual} by re-annotating its images with $53$ ISP-relevant quality issues using GPT-5, creating a benchmark for structured issue-level diagnosis. Second, we validate the approach on a small-scale expert-annotated real-world dataset, demonstrating that PrISM-IQA remains effective beyond the reconstructed public benchmark. In addition, we assess whether the learned representations capture broader perceptual information by using linear probing for MOS prediction. The strong probing results suggest that PrISM-IQA learns rich and transferable quality representations rather than narrowly fitting relabeled issue tags. Finally, using OpenISP~\footnote{https://github.com/cruxopen/openISP} as a controllable simulated ISP testbed, we show in Appendix~\ref{app:openisp_tuning} that the structured predictions of PrISM-IQA can be translated into actionable feedback for ISP tuning.

In summary, our main contributions are threefold.

\begin{itemize}
    \item We introduce PrISM-IQA, a practical reformulation of smartphone IQA that replaces scalar MOS prediction with multi-issue ordinal diagnosis better aligned with ISP tuning workflows.
    \item We develop a structured prediction layer that combines cumulative ordinal encoding and structured inference to model ordinal labels and their semantic dependencies.
    \item We validate PrISM-IQA on reconstructed SPAQ and expert-annotated real-world data, while further showing the perceptual relevance and practical value of the learned representations.
\end{itemize}

\section{Related Work}
In this section, we review existing work that is most relevant to our study: no reference IQA, IQA datasets especially for smartphone photography, and structured prediction with ordinal labels. 

\subsection{No-Reference IQA Models}
Early no-reference IQA methods were largely developed for synthetic distortions and often relied on natural scene statistics, as exemplified by BRISQUE~\cite{mittal2012no} and NIQE~\cite{mittal2013niqe}. While influential, such methods are known to struggle on authentically distorted images in the wild, where degradations arise from coupled effects of scene content, optics, sensors, and in-camera processing rather than from a single controlled corruption. Subsequent work therefore shifted toward NR-IQA for authentic, real-world distortions, including deep bilinear pooling for mixed distortion scenarios~\cite{zhang2018blind}, content-adaptive regression~\cite{su2020blindly}, attention-based modeling such as MUSIQ~\cite{ke2021musiq} and TOPIQ~\cite{chen2024topiq},  vision-language correspondence such as LIQE~\cite{zhang2023blind}, and more recently, reasoning-induced prediction via reinforcement learning like Q-Insight~\cite{li2025qinsight} and VisualQuality-R1~\cite{wu2025visualqualityr1}. Despite their strong predictive performance, these methods still primarily treat no-reference IQA as scalar MOS regression or ranking. As a result, they provide limited insight into why an image is judged poor, which specific artifacts are responsible, and how severe each issue is. This limitation reduces their usefulness for smartphone ISP development, where practitioners need interpretable, issue-level feedback rather than a single overall quality estimate.

\subsection{IQA Datasets for Authentic Distortions}
Progress on NR-IQA for authentic distortions has been driven by datasets collected in the wild~\cite{virtanen2014cid2013}. LIVE Challenge~\cite{ghadiyaram2015massive} established a large-scale benchmark for photographic distortions, while KonIQ-10k~\cite{hosu2020koniq}, PaQ-2-PiQ~\cite{ying2020paq2piq}, and UHD-IQA~\cite{hosu2024uhd} further expanded the scale, resolution, and ecological validity. For smartphone photography in particular, SPAQ~\cite{fang2020perceptual} is especially relevant because it targets smartphone-captured images and augments MOSs with five continuous perceptual attributes and scene labels. SQAD~\cite{fang2023sqad} complements this direction by benchmarking device-level smartphone camera quality. These resources are highly valuable, but their supervision remains relatively coarse for ISP diagnosis: they emphasize overall quality or a small number of continuous attributes, rather than fine-grained issue identities with ordered severities and semantic dependencies. In this work, we reconstruct SPAQ by enriching it with fine-grained issue categories and ordered severity annotations tailored to smartphone ISP diagnosis, and we further complement it with a small-scale, expert-annotated, real-world dataset for additional validation.

\subsection{Ordinal and Structured Prediction in Vision and Learning}
Our formulation is also related to the broader literature on ordinal prediction in machine learning. Prior work has modeled ordinal structure through ranking-based formulations~\cite{herbrich2000large,crammer2001pranking} that learn relative order between labels, threshold-based or cumulative approaches~\cite{mccullagh1980regression,chu2007support} that predict whether a label exceeds a sequence of ordered decision boundaries, as well as reduction-based methods~\cite{frank2001simple,li2006ordinal} that further cast a $K$-level ordinal problem into $K-1$ binary subtasks. In computer vision, such formulations have been widely used in applications such as age estimation~\cite{li2012learning,niu2016ordinal}, monocular depth estimation~\cite{fu2018deep}, and medical disease grading~\cite{liu2018ordinal}, where modeling and respecting label order improve prediction accuracy.

A parallel line of work studies structured prediction, where outputs are not assumed to be independent~\cite{nowozin2011structured}. 
General frameworks such as conditional random fields~\cite{lafferty2001conditional} and structured large-margin methods~\cite{tsochantaridis2005large} are now standard for constraint-aware vision outputs, spanning scene labeling~\cite{he2004multiscale}, human pose estimation~\cite{felzenszwalb2005pictorial}, stereo and multiview reconstruction~\cite{kolmogorov2001computing}, and joint object detection and segmentation~\cite{gould2009region}. Particularly relevant here are label-relation models~\cite{deng2014large} that encode semantic structure directly, such as hierarchy and exclusion constraints among labels. These ideas are a natural fit for ISP issue diagnosis. In our setting, each quality issue is associated with an ordered severity level, while different issues may also obey cross-label constraints such as subsumption and exclusion. 

\section{Proposed Method: PrISM-IQA}
\begin{figure}
    \centering
    \includegraphics[width=\linewidth]{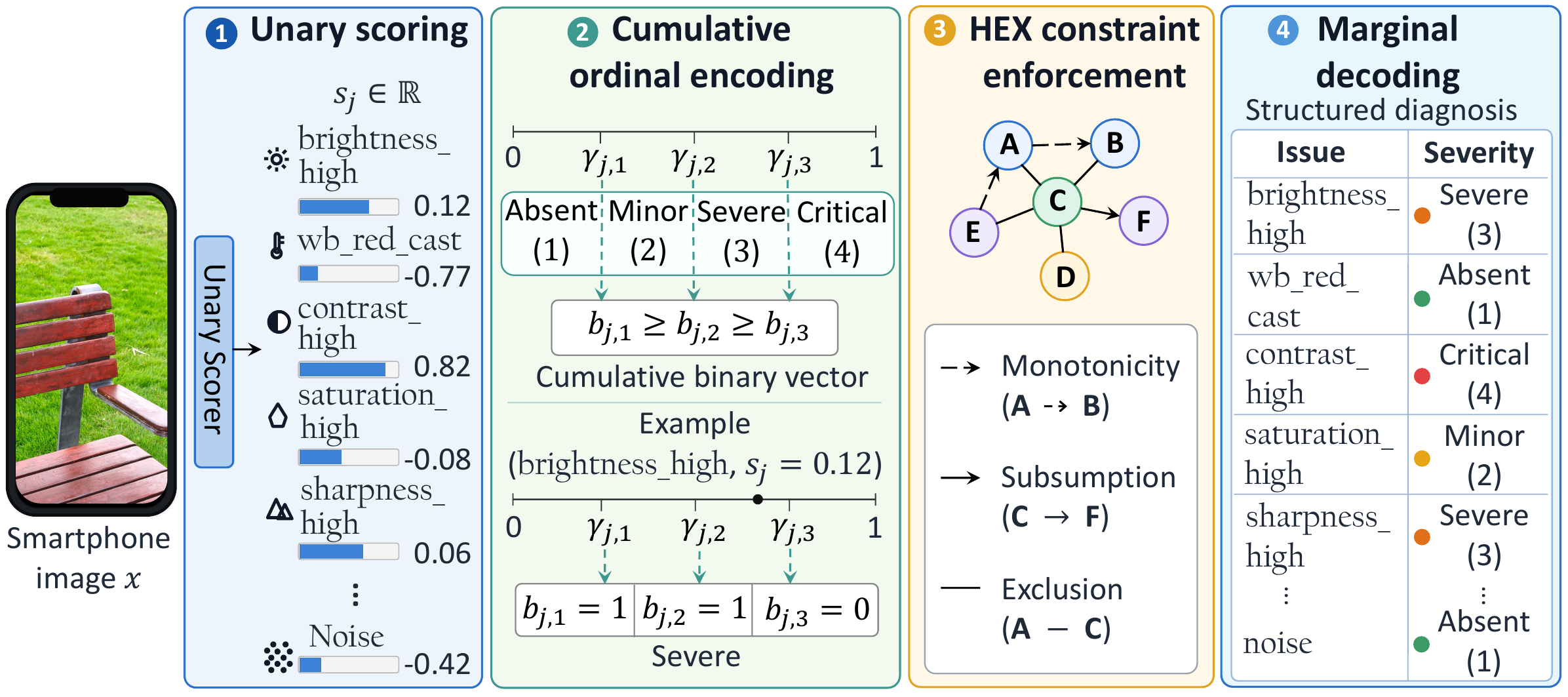}
    \caption{Overview of PrISM-IQA. Given a smartphone image \(x\), the model computes a unary severity score \(s_j(x)\) for each quality issue \(j\), converts ordinal labels into cumulative events \(b_{j,k}=\mathbb{I}[y_j>k]\), and performs HEX-constrained multi-issue ordinal inference over \(b=\{\{b_{j,k}\}_{k=1}^{K_j-1}\}_{j=1}^{N}\). The HEX layer enforces within-issue monotonicity together with subsumption and exclusion relations; marginal decoding then returns logically consistent issue-level severity diagnosis.}
    \label{fig:model}
\end{figure}

In this section, we present the proposed PrISM-IQA in detail, which has two key components. First, we represent each quality issue by a cumulative ordinal encoding and impose semantic relations among issues through a hierarchy-and-exclusion (HEX) graph~\cite{deng2014large}. Second, we instantiate the resulting structured predictor with a vision-language model~\cite{zhang2023blind} that produces issue-aware unary scores from image-text correspondence. The system diagram of PrISM-IQA is shown in Fig.~\ref{fig:model}.

\subsection{Problem Formulation}
We formulate smartphone IQA as a multi-issue ordinal diagnosis problem. Let
\begin{equation}
f:\mathcal{X}\mapsto\mathcal{Y}, \qquad 
\mathcal{Y}=\prod_{j=1}^{N}\mathcal{Y}_j, \qquad 
\mathcal{Y}_j=\{1,\ldots,K_j\},
\end{equation}
be a smartphone IQA system, where $\mathcal{X}$ denotes the image space, $N$ is the number of quality issues, and $y_j\in\mathcal{Y}_j$ denotes the severity level of issue $j$. Larger label values correspond to more severe degradation. In our setting, $K_j=4$, corresponding to \emph{absent}, \emph{minor}, \emph{severe}, and \emph{critical} levels, respectively, although the formulation is not restricted to the same number of levels for all issues. Given an input image $x\in\mathcal{X}$, the quality diagnosis system predicts an issue-wise severity vector
$f(x)=[\hat{y}_1,\ldots,\hat{y}_N]\in\mathcal{Y}$ rather than a single overall quality score.

The training set consists of $M$ partially annotated samples of the form $
\mathcal{D}=\{(x^{(i)}, \{y_j^{(i)}\}_{j\in\Omega^{(i)}})\}_{i=1}^{M}$,
where $\Omega^{(i)}\subseteq\{1,\ldots,N\}$ denotes the subset of quality issues annotated for image $x^{(i)}$. This explicitly allows missing labels, since not every issue must be labeled for every training image in the real world. A practical smartphone IQA system should satisfy several requirements. During training, it must be able to learn from incomplete supervision and capture semantic dependencies among quality issues. During inference, it must produce structured predictions that both respect the ordinal nature of severity levels within each issue and remain globally consistent with cross-issue semantic relations, such as subsumption and mutual exclusion. The ultimate objective is to learn a mapping from an input image to complete, interpretable, and logically coherent multi-issue quality diagnosis.

\subsection{Cumulative Encoding for Ordinal Labels}\label{subsec:cumenc}
To model the ordinal structure of issue severity, we convert each issue label $y_j$ into $K_j-1$ cumulative binary variables:
\begin{equation}
b_{j,k} = \mathbb{I}[y_j > k], \qquad k = 1, \ldots, K_j - 1,
\end{equation}
where $\mathbb{I}[\cdot]$ denotes the indicator function. Each binary variable $b_{j,k}$ indicates whether the severity of issue $j$ exceeds threshold $k$. This encoding naturally induces the monotonicity property:
$b_{j,k+1} \leq b_{j,k}$, for  $k = 1, \ldots, K_j - 2$,
since exceeding a higher threshold implies exceeding all lower thresholds. For example, when $K_j = 4$: $y_j = 1$ (\ie, \emph{absent}) implies $[b_{j,1}, b_{j,2}, b_{j,3}] = [0,0,0]$; $y_j = 2$ (\ie, \emph{minor}) implies $[1,0,0]$; $y_j = 3$ (\ie, \emph{severe}) implies $[1,1,0]$; and $y_j = 4$ (\ie, \emph{critical}) implies $[1,1,1]$.
The collection
$b = \{\{b_{j,k}\}_{k=1}^{K_j-1}\}_{j=1}^{N}$
forms the concatenated binary vector, whose total dimensionality is
$K = \sum_{j=1}^{N}(K_j - 1)$. 

\subsection{HEX Graph for Issue Relations}\label{subsec:hex}
After the cumulative transformation, we construct a HEX graph
$\mathcal{G} = (\mathcal{V}, \mathcal{E}_h, \mathcal{E}_e)$
over the binary variables. Each node $v_{j,k} \in \mathcal{V}$ corresponds to the binary event $b_{j,k}=1$, \ie, that the severity of issue $j$ exceeds threshold $k$. The graph contains two types of edges: hierarchy edges $\mathcal{E}_h$, which encode implication constraints, and exclusion edges $\mathcal{E}_e$, which encode mutually exclusive events. The hierarchy edges include both within-issue monotonicity and cross-issue subsumption relations.

\noindent\textbf{Within-Issue Monotonicity.}
For each issue $j$ and threshold $k \in \{1,\dots,K_j-2\}$, we add a directed edge
$v_{j,k+1} \rightarrow v_{j,k}$. Its hard constraint potential is defined as
\begin{equation}\label{eq:mono}
\psi^{\mathrm{hier}}_{(j,k+1)\rightarrow(j,k)}(b_{j,k+1}, b_{j,k})
=
\mathbb{I}\!\left[b_{j,k+1} \leq b_{j,k}\right],
\end{equation}
which excludes the invalid configuration $[b_{j,k+1}, b_{j,k}] = [1,0]$.

\noindent\textbf{Cross-Issue Subsumption.} 
If one event (\eg, $b_{j,k} = 1$) implies another ($b_{j',k'} = 1$), for example, a specific defect implies the presence of a broader parent defect, we add a directed hierarchy edge $v_{j,k} \rightarrow v_{j',k'}$, with
the hard constraint potential defined as
\begin{equation}\label{eq:sub}
\psi^{\mathrm{hier}}_{(j,k)\rightarrow(j',k')}(b_{j,k}, b_{j',k'})
=
\mathbb{I}\!\left[b_{j,k} \leq b_{j',k'}\right].
\end{equation}
This rules out the inconsistent configuration $[b_{j,k}, b_{j',k'}] = [1,0]$.

\noindent\textbf{Cross-Issue Exclusion.} If two cumulative events cannot occur simultaneously, for example, ``\texttt{brightness\_high}'' and ``\texttt{brightness\_low}'' at incompatible severity levels, we add an undirected exclusion edge $
v_{j,k} - v_{j',k'}$. Its hard constraint potential is defined as
\begin{equation}\label{eq:excl}
\psi^{\mathrm{excl}}_{(j,k),(j',k')}(b_{j,k}, b_{j',k'})
=
\mathbb{I}\!\left[b_{j,k} + b_{j',k'} \leq 1\right].
\end{equation}

\subsection{HEX-Constrained Multi-Issue Ordinal Inference}\label{subsec:stm}
Given the cumulative encoding in Sec.~\ref{subsec:cumenc} and the HEX graph in Sec.~\ref{subsec:hex}, we formulate smartphone IQA as structured prediction over the binary variables $
b=\{b_{j,k}\}$.
Recall that each node $v_{j,k}\in \mathcal{V}$ corresponds to the cumulative event $b_{j,k}=1$, namely that issue \(j\) exceeds severity threshold \(k\). To preserve the ordinal semantics of the cumulative construction, we associate one ``raw'' severity score with each issue, rather than one independent score with each cumulative node. Specifically, let
$s_j(x)\in\mathbb{R}$ 
denote the image-dependent raw severity score for issue \(j\), and let
$\gamma_{j,1}<\gamma_{j,2}<\cdots<\gamma_{j,K_j-1}$ be ordered thresholds for that issue. We define the unary log-potential for node \(v_{j,k}\) as
\begin{equation}\label{eq:orderedrep}
\bar{s}_{j,k}(x)=s_j(x)-\gamma_{j,k},
\qquad k=1,\ldots,K_j-1 ,\;\; j=1,\ldots,N.
\end{equation}
Since all cumulative events of issue \(j\) are generated from the same raw score \(s_j(x)\) and differ only through ordered thresholds, the unary model itself admits a valid ordinal interpretation: larger \(k\) corresponds to a stricter exceedance condition. Alternatively, one may assign an independent score \(s_{j,k}(x)\) to each cumulative node \(v_{j,k}\). Although this node-wise parameterization is more expressive, it does not by itself enforce monotonicity. Nonetheless, the hard HEX constraints in Eqs.~\eqref{eq:mono}-\eqref{eq:excl} can resolve this issue at the joint-inference stage by excluding invalid label configurations, so the resulting prediction remains structurally consistent. Empirically, the node-wise variant can achieve similar performance when combined with HEX-constrained inference, but Eq.~\eqref{eq:orderedrep} offers a more principled parameterization for ordinal modeling and is therefore adopted as the default. Consequently, the unary potential is defined as \begin{equation}
\phi_{j,k}(b_{j,k}\mid x)
=
\exp\!\left(\bar{s}_{j,k}(x)\,b_{j,k}\right)
=
\exp\!\left((s_j(x)-\gamma_{j,k})\,b_{j,k}\right).
\end{equation}

Using these unary potentials together with the HEX constraints, we define the following unnormalized conditional distribution:
\begin{equation}
\tilde{P}(b\mid x)
=
\left(
\prod_{j=1}^{N}\prod_{k=1}^{K_j-1}
\phi_{j,k}(b_{j,k}\mid x)
\right)
\left(
\prod_{u\to v \in \mathcal{E}_h}
\psi^{\mathrm{hier}}_{u\to v}(b_u,b_v)
\right)
\left(
\prod_{(u,v)\in \mathcal{E}_e}
\psi^{\mathrm{excl}}_{u,v}(b_u,b_v)
\right).
\end{equation}
Thus, the unary terms provide image-dependent evidence, while the HEX graph restricts inference to assignments that are both ordinally and semantically valid. The conditional random field distribution is then obtained by normalizing \(\tilde{P}(b \mid x)\):
\begin{equation}
P(b \mid x)
=
\frac{\tilde{P}(b \mid x)}{Z(x)},
\end{equation}
where the partition function is defined as $Z(x)=\sum_{b' \in \{0,1\}^{K}} \tilde{P}(b' \mid x)$. 

In practice, exact inference over the full joint state space becomes expensive when the number of issue labels is large. We therefore exploit the sparsity of the HEX constraints and factorize the graph into disconnected components, performing exact inference independently within each component. This preserves exactness while making inference tractable in the large-label setting.

From this distribution, the issue-wise ordinal probabilities can be recovered by marginalization as
\begin{equation}
P(y_j = k \mid x)
=
P(b_{j,k-1}=1 \mid x) - P(b_{j,k}=1 \mid x),
\qquad k = 1, \dots, K_j,
\end{equation}
where \(P(b_{j,0}=1 \mid x)=1\) and \(P(b_{j,K_j}=1 \mid x)=0\).
The final prediction for issue \(j\) is obtained by maximum a posteriori decoding:
\begin{equation}\label{eq:map}
\hat{y}_j
=
\argmax_{k \in \{1,\ldots,K_j\}} P(y_j = k \mid x).
\end{equation}

Training is performed by minimizing the negative marginal log-likelihood of the observed labels under the joint model:
\begin{equation}\label{eq:finalloss}
\ell(\mathcal{B})
=
-\sum_{(x,y)\in\mathcal{B}}
\log P(y_{\Omega}\mid x)
=
-\sum_{(x,y)\in\mathcal{B}}
\log
\sum_{y_{\Omega^\complement}}
P(y_{\Omega}, y_{{\Omega^\complement}} \mid x),
\end{equation}
where \(\mathcal{B}\) denotes a minibatch of training examples, \(\Omega\) is the set of observed issue indices for a given sample, \(y_{\Omega}=\{y_j\}_{j\in\Omega}\) are the corresponding observed issue labels, and \(y_{{\Omega^\complement}}\) denotes the unobserved issue labels on the complementary index set \({\Omega^\complement}\). Thus, Eq.~\eqref{eq:finalloss} allows the model to be trained from partially annotated data while still learning the full joint structured distribution over all issues.

\subsection{Vision-Language Instantiation}
The structured formulation in Secs.~\ref{subsec:cumenc}-\ref{subsec:stm} is agnostic to the choice of unary scorer. Here, we instantiate it with CLIP~\cite{radford2021learning}, the vision-language backbone underlying the state-of-the-art no-reference IQA model LIQE~\cite{zhang2023blind}. Consistent with Eq.~\eqref{eq:orderedrep}, the goal of this module is to produce, for each quality issue $j$, a single raw severity score $s_j(x)$ from which the cumulative node scores are derived through issue-specific ordered thresholds.

Let \(e: \mathbb{R}^{H \times W \times 3} \rightarrow \mathbb{R}^{D\times 1}\) and \(
g : \mathcal{T} \rightarrow \mathbb{R}^{D\times 1}\)
denote the image encoder and text encoder, respectively, where both modalities are mapped into the same $D$-dimensional embedding space. Given an input image $x$, we first compute the normalized visual embedding
\(
\bar{e}(x) = \frac{e(x)}{\|e(x)\|_2}.
\) For each issue $j \in \{1,\dots,N\}$, let $\mathrm{text}_j$ denote its textual description. Rather than using a handcrafted prompt template, we adopt the CoOp strategy~\cite{zhou2022learning} and prepend a learnable context to the issue tokens. Specifically, let $c = [c_1,\ldots,c_L]$ be a sequence of $L$ learnable context tokens shared across issues. The prompt associated with issue $j$ is then \(
t_j = [c, \mathrm{Tok}(\mathrm{text}_j)] \),
where $\mathrm{Tok}(\cdot)$ denotes the tokenizer of the text encoder. The corresponding normalized textual embedding is \(
\bar{g}_j = \frac{g(t_j)}{\|g(t_j)\|_2}\). 

We define the raw severity score of issue $j$ for image $x$ as the scaled cosine similarity between the visual and textual embeddings:
\begin{equation}
    s_j(x) = \tau\, 	\langle\bar{e}(x), \bar{g}_j	\rangle,
\end{equation}
where $\tau > 0$ is a learnable temperature parameter. 

This design makes the vision-language component and the HEX-constrained inference complementary. The vision-language backbone produces semantically grounded, issue-level severity evidence, while the structured layer in Sec.~\ref{subsec:stm} converts these scores into ordinally valid and semantically consistent multi-issue predictions by enforcing within-issue monotonicity and cross-issue relations during joint inference.  This yields a modular diagnosis pipeline: the unary scorer can be replaced by a stronger visual backbone or a different prompting strategy, while the structured inference remains unchanged.

\section{Experiments}
\label{sec:exp}
We evaluate PrISM-IQA from three complementary perspectives: 1) whether the proposed issue taxonomy and structured inference yield meaningful diagnosis on a large public benchmark, 2) whether the resulting model generalizes to real-world expert annotations collected in an industrial smartphone photography setting, and 3) whether the learned representation retains broader perceptual information beyond the issue labels used for training.

\subsection{Experimental Protocols}
\noindent\textbf{Datasets.}
We use two datasets. The first is a reconstructed version of SPAQ~\cite{fang2020perceptual} with fine-grained multi-issue ordinal labels, which serves as the main benchmark for large-scale controlled evaluation. The second is an expert-annotated dataset of $2,983$ smartphone photographs, which serves as a real-world validation set. Further details on dataset construction are provided in Appendix~\ref{app:datasets}.

\definecolor{overallbg}{RGB}{255,249,238}
\definecolor{facebg}{RGB}{252,242,244}
\definecolor{buildingbg}{RGB}{242,246,255}
\definecolor{skybg}{RGB}{240,250,255}
\definecolor{greenerybg}{RGB}{242,250,242}
\definecolor{maincolorbg}{RGB}{255,246,239}
\definecolor{summarybg}{RGB}{245,245,245}
\begin{table}[t]
\centering
\footnotesize
\setlength{\tabcolsep}{3pt}
\caption{Issue-wise performance of PrISM-IQA on reconstructed SPAQ.  \texttt{global} denotes global image-level issues, while \texttt{building}, \texttt{face},  \texttt{greenery}, \texttt{sky}, and \texttt{dominant\_color} denote local content-dependent issues. ACC, tAUC, and QWK are reported as percentages (\%). Each category block ends with its mean, and the final row reports the mean over all evaluated issues.}
\begin{minipage}[t]{0.47\textwidth}
\vspace{0pt}
\centering
\begin{tabular}{@{}>{\raggedright\arraybackslash}p{0.18\linewidth}
                >{\raggedright\arraybackslash}p{0.42\linewidth}
                >{\centering\arraybackslash}p{0.10\linewidth}
                >{\centering\arraybackslash}p{0.10\linewidth}
                >{\centering\arraybackslash}p{0.10\linewidth}@{}}
\toprule
Scope & Issue & ACC$\uparrow$ & tAUC$\uparrow$ & QWK$\uparrow$ \\
\midrule

\rowcolor{buildingbg}
\texttt{building} & \texttt{brightness\_high} & 93.55 & 95.42 & 61.58 \\
\rowcolor{buildingbg}
& \texttt{brightness\_low} & 92.36 & 96.94 & 70.78 \\
\rowcolor{buildingbg}
& \texttt{clarity\_low} & 60.34 & 87.52 & 74.80 \\
\rowcolor{buildingbg}
& \texttt{contrast\_low} & 97.70 & 92.40 & 28.38 \\
\rowcolor{buildingbg}
& \textbf{mean} & \textbf{85.99} & \textbf{93.07} & \textbf{58.88} \\

\addlinespace[2pt]

\rowcolor{facebg}
\texttt{face} & \texttt{brightness\_high} & 96.31 & 95.30 & 32.69 \\
\rowcolor{facebg}
& \texttt{brightness\_low} & 98.96 & 97.29 & 46.62 \\
\rowcolor{facebg}
& \texttt{clarity\_low} & 90.42 & 95.78 & 69.00 \\
\rowcolor{facebg}
& \texttt{contrast\_high} & 96.82 & 95.82 & 30.15 \\
\rowcolor{facebg}
& \texttt{contrast\_low} & 98.93 & 93.69 & 5.87 \\
\rowcolor{facebg}
& \texttt{hair\_detail\_loss} & 95.94 & 97.17 & 55.66 \\
\rowcolor{facebg}
& \texttt{noise} & 95.44 & 96.10 & 53.55 \\
\rowcolor{facebg}
& \texttt{saturation\_low} & 97.03 & 96.02 & 30.96 \\
\rowcolor{facebg}
& \texttt{texture\_artifact} & 98.57 & 96.80 & 11.35 \\
\rowcolor{facebg}
& \textbf{mean} & \textbf{96.49} & \textbf{96.00} & \textbf{37.31} \\

\addlinespace[2pt]

\rowcolor{greenerybg}
\texttt{greenery} & \texttt{clarity\_low} & 66.93 & 89.70 & 77.58 \\
\rowcolor{greenerybg}
& \texttt{contrast\_high} & 98.83 & 86.94 & 25.63 \\
\rowcolor{greenerybg}
& \texttt{contrast\_low} & 65.58 & 85.17 & 58.41 \\
\rowcolor{greenerybg}
& \texttt{saturation\_high} & 95.30 & 86.57 & 30.19 \\
\rowcolor{greenerybg}
& \texttt{saturation\_low} & 82.11 & 88.94 & 63.17 \\
\rowcolor{greenerybg}
& \textbf{mean} & \textbf{81.75} & \textbf{87.46} & \textbf{51.00} \\

\bottomrule
\end{tabular}

\end{minipage}
\hspace{0.03\textwidth}
\begin{minipage}[t]{0.47\textwidth}
\vspace{0pt}
\centering
\begin{tabular}{@{}>{\raggedright\arraybackslash}p{0.18\linewidth}
                >{\raggedright\arraybackslash}p{0.42\linewidth}
                >{\centering\arraybackslash}p{0.10\linewidth}
                >{\centering\arraybackslash}p{0.10\linewidth}
                >{\centering\arraybackslash}p{0.10\linewidth}@{}}
\toprule
Scope & Issue & ACC$\uparrow$ & tAUC$\uparrow$ & QWK$\uparrow$ \\
\midrule

\rowcolor{skybg}
\texttt{sky} & \texttt{contrast\_low} & 68.76 & 85.49 & 63.57 \\
\rowcolor{skybg}
& \texttt{highlight\_clipped} & 82.64 & 87.92 & 76.67 \\
\rowcolor{skybg}
& \texttt{highlight\_muted} & 96.87 & 86.30 & 13.86 \\
\rowcolor{skybg}
& \texttt{saturation\_low} & 79.35 & 91.41 & 75.79 \\
\rowcolor{skybg}
& \textbf{mean} & \textbf{81.91} & \textbf{87.78} & \textbf{57.47} \\

\addlinespace[2pt]

\rowcolor{maincolorbg}
 & \texttt{color\_cast} & 76.85 & 90.16 & 59.07 \\
\rowcolor{maincolorbg}
& \texttt{saturation\_high} & 59.33 & 80.52 & 34.39 \\
\rowcolor{maincolorbg}
& \texttt{saturation\_low} & 78.75 & 59.17 & 8.04 \\
\rowcolor{maincolorbg}
\multirow[t]{-4}{=}{\makecell[tl]{\texttt{dominant\_}\\\texttt{color}}} & \textbf{mean} & \textbf{71.64} & \textbf{76.61} & \textbf{33.83} \\

\addlinespace[2pt]

\rowcolor{overallbg}
\texttt{global} & \texttt{brightness\_high} & 91.78 & 94.77 & 79.78 \\
\rowcolor{overallbg}
& \texttt{brightness\_low} & 79.25 & 91.78 & 58.97 \\
\rowcolor{overallbg}
& \texttt{clarity\_low} & 68.18 & 86.67 & 60.08 \\
\rowcolor{overallbg}
& \texttt{contrast\_high} & 93.53 & 83.92 & 44.16 \\
\rowcolor{overallbg}
& \texttt{contrast\_low} & 66.65 & 71.68 & 29.59 \\
\rowcolor{overallbg}
& \texttt{highlight\_clipped} & 78.46 & 92.36 & 70.23 \\
\rowcolor{overallbg}
& \texttt{noise} & 86.99 & 92.74 & 60.17 \\
\rowcolor{overallbg}
& \texttt{wb\_blue\_cast} & 96.36 & 82.88 & 16.95 \\
\rowcolor{overallbg}
& \texttt{wb\_yellow\_cast} & 91.74 & 84.38 & 29.99 \\
\rowcolor{overallbg}
& \textbf{mean} & \textbf{83.66} & \textbf{86.80} & \textbf{49.99} \\

\addlinespace[2pt]

\rowcolor{summarybg}
\textbf{mean} & & \textbf{85.78} & \textbf{89.29} & \textbf{47.28} \\
\bottomrule
\end{tabular}
\end{minipage}
\label{tab:local_feature_merged}

\end{table}

\begin{table}[t]
\centering
\caption{Issue-wise performance of PrISM-IQA on the expert-annotated dataset.}
\label{tab:expert_results}
\footnotesize
\renewcommand{\arraystretch}{1.08}

\begin{tabular}{lccc|lccc}
\toprule
Issue & ACC$\uparrow$ & tAUC$\uparrow$ & QWK$\uparrow$
& Issue & ACC$\uparrow$ & tAUC$\uparrow$ & QWK$\uparrow$ \\
\midrule
\texttt{brightness\_high} 
& 92.98 & 85.01 & 35.45
& \texttt{wb\_blue\_cast} 
& 79.26 & 87.64 & 60.64 \\

\texttt{brightness\_low} 
& 89.30 & 72.42 & 13.66
& \texttt{wb\_red\_cast} 
& 92.31 & 91.81 & 67.68 \\

\texttt{contrast\_high} 
& 94.98 & 87.85 & 17.80
& \texttt{wb\_yellow\_cast} 
& 85.95 & 85.12 & 48.89 \\

\texttt{contrast\_low} 
& 89.63 & 88.43 & 29.62
& \textbf{mean} 
& \textbf{89.20} & \textbf{85.47} & \textbf{39.11} \\
\bottomrule
\end{tabular}
\end{table}

\noindent\textbf{Implementation Details.}
We adopt ViT-L/14~\cite{radford2021learning} as the visual encoder and GPT-2~\cite{radford2019language} as the text encoder. The model is optimized with AdamW~\cite{loshchilov2017decoupled} using a weight decay of $10^{-3}$. The initial learning rates of the visual encoder and CoOp are set to $5\times10^{-6}$ and $5\times10^{-4}$, respectively, and scheduled with cosine annealing~\cite{loshchilov2016sgdr}. We use $L=4$ learnable context tokens in CoOp. Training runs for 50 epochs with a mini-batch size of 50. During training and inference, we randomly crop $5$ and $9$ sub-images as in LIQE~\cite{zhang2023blind} to compute the mean image-level corresponding score, respectively, each of spatial size $224\times224\times3$. 

\noindent\textbf{Competing Methods.}
For issue-level diagnosis, we compare PrISM-IQA against representative learning-based NR-IQA models, including MUSIQ~\cite{ke2021musiq}, DBCNN~\cite{zhang2018blind}, TReS~\cite{golestaneh2022no}, UNIQUE~\cite{zhang2021uncertainty}, LIQE~\cite{zhang2023blind}, TOPIQ~\cite{chen2024topiq}, and
Q-Insight~\cite{li2025qinsight}. 
All baselines are adapted to our issue-level ordinal label space and trained on the same data splits. For standard NR-IQA models, we replace the original scalar quality head with an issue-level unary scoring head. Q-Insight is adapted differently: rather than reproducing its original reasoning-induced pipeline, we use its backbone with a fixed image–issue prompt and supervised ordinal labels. Each predefined issue has a severity token whose hidden representation is projected to a unary score. 

\definecolor{crfbg}{RGB}{244,249,255}
\definecolor{impgray}{RGB}{120,120,120}
\newcommand{\crfcell}[1]{\cellcolor{crfbg}#1}
\newcommand{\best}[1]{\textbf{#1}}
\newcommand{\second}[1]{\underline{#1}}
\newcommand{\impcell}[1]{\textcolor{impgray}{\scriptsize #1\%}}
\newcommand{\improw}{\textcolor{impgray}{\scriptsize \emph{Improvement}}}
\begin{table}[t]
\centering
\scriptsize
\setlength{\tabcolsep}{2.2pt}
\renewcommand{\arraystretch}{1.10}
\caption{Effect of PrISM-based structured inference across NR-IQA backbones on reconstructed SPAQ. Each backbone is adapted as an issue-level unary scorer, with and without PrISM, isolating the contribution of cumulative ordinal encoding and HEX-constrained multi-issue inference. Results report category-wise and mean tAUC and QWK in percentages. \emph{Improvement} rows show the relative percentage change from w/o PrISM to w/ PrISM.}
\label{tab:main_comparison}

\resizebox{\textwidth}{!}{%
\begin{tabular}{l|l|cc|cc|cc|cc|cc|cc|cc}
\toprule
\multirow{2}{*}{Method} 
& \multirow{2}{*}{Variant}
& \multicolumn{2}{c|}{\texttt{building}}
& \multicolumn{2}{c|}{\texttt{face}}
& \multicolumn{2}{c|}{\texttt{greenery}}
& \multicolumn{2}{c|}{\texttt{sky}}
& \multicolumn{2}{c|}{\texttt{dominant\_color}}
& \multicolumn{2}{c|}{\texttt{global}}
& \multicolumn{2}{c}{mean} \\
\cmidrule(lr){3-4}
\cmidrule(lr){5-6}
\cmidrule(lr){7-8}
\cmidrule(lr){9-10}
\cmidrule(lr){11-12}
\cmidrule(lr){13-14}
\cmidrule(lr){15-16}
& 
& tAUC$\uparrow$ & QWK$\uparrow$
& tAUC$\uparrow$ & QWK$\uparrow$
& tAUC$\uparrow$ & QWK$\uparrow$
& tAUC$\uparrow$ & QWK$\uparrow$
& tAUC$\uparrow$ & QWK$\uparrow$
& tAUC$\uparrow$ & QWK$\uparrow$
& tAUC$\uparrow$ & QWK$\uparrow$ \\
\midrule

\multirow{3}{*}{MUSIQ~\cite{ke2021musiq}}
& w/o PrISM
& 71.93 & 40.83
& 60.14 & 2.44
& 69.01 & 37.63
& 74.16 & 45.68
& 62.39 & 27.28
& 76.13 & 41.11
& 68.91 & 29.65 \\

& \crfcell{w/ PrISM}
& \crfcell{91.44} & \crfcell{42.55}
& \crfcell{74.13} & \crfcell{7.77}
& \crfcell{84.75} & \crfcell{34.97}
& \crfcell{85.45} & \crfcell{40.66}
& \crfcell{{78.37}} & \crfcell{27.14}
& \crfcell{86.19} & \crfcell{44.11}
& \crfcell{82.63} & \crfcell{31.06} \\

& \improw
& \impcell{+27.1} & \impcell{+4.2}
& \impcell{+23.3} & \impcell{+218.4}
& \impcell{+22.8} & \impcell{-7.1}
& \impcell{+15.2} & \impcell{-11.0}
& \impcell{+25.6} & \impcell{-0.5}
& \impcell{+13.2} & \impcell{+7.3}
& \impcell{+19.9} & \impcell{+4.8} \\
\midrule

\multirow{3}{*}{DBCNN~\cite{zhang2018blind}}
& w/o PrISM
& 93.60 & {58.64}
& 85.38 & {43.01}
& 82.61 & 36.68
& 75.55 & 36.78
& 77.35 & 29.16
& 84.72 & 44.84
& 83.90 & 42.45 \\

& \crfcell{w/ PrISM}
& \crfcell{{95.27}} & \crfcell{56.35}
& \crfcell{86.70} & \crfcell{{46.40}}
& \crfcell{{85.61}} & \crfcell{34.31}
& \crfcell{80.76} & \crfcell{42.27}
& \crfcell{76.99} & \crfcell{29.93}
& \crfcell{{87.41}} & \crfcell{46.72}
& \crfcell{86.18} & \crfcell{43.94} \\

& \improw
& \impcell{+1.8} & \impcell{-3.9}
& \impcell{+1.5} & \impcell{+7.9}
& \impcell{+3.6} & \impcell{-6.5}
& \impcell{+6.9} & \impcell{+14.9}
& \impcell{-0.5} & \impcell{+2.6}
& \impcell{+3.2} & \impcell{+4.2}
& \impcell{+2.7} & \impcell{+3.5} \\
\midrule

\multirow{3}{*}{TReS~\cite{golestaneh2022no}}
& w/o PrISM
& 78.48 & 36.76
& 76.88 & 21.71
& 75.14 & 36.34
& 70.88 & 35.95
& 63.76 & 20.03
& 74.16 & 37.02
& 74.19 & 31.21 \\

& \crfcell{w/ PrISM}
& \crfcell{90.32} & \crfcell{42.56}
& \crfcell{78.43} & \crfcell{12.09}
& \crfcell{82.12} & \crfcell{35.04}
& \crfcell{79.92} & \crfcell{40.91}
& \crfcell{73.87} & \crfcell{25.53}
& \crfcell{83.55} & \crfcell{41.39}
& \crfcell{81.50} & \crfcell{31.38} \\

& \improw
& \impcell{+15.1} & \impcell{+15.8}
& \impcell{+2.0} & \impcell{-44.3}
& \impcell{+9.3} & \impcell{-3.6}
& \impcell{+12.8} & \impcell{+13.8}
& \impcell{+15.9} & \impcell{+27.5}
& \impcell{+12.7} & \impcell{+11.8}
& \impcell{+9.9} & \impcell{+0.5} \\
\midrule

\multirow{3}{*}{UNIQUE~\cite{zhang2021uncertainty}}
& w/o PrISM
& 84.37 & 34.67
& 82.47 & 14.41
& 76.56 & 37.62
& 80.75 & 40.76
& 71.06 & 26.37
& 76.03 & 35.79
& 78.91 & 30.02 \\

& \crfcell{w/ PrISM}
& \crfcell{{94.01}} & \crfcell{53.18}
& \crfcell{81.18} & \crfcell{32.59}
& \crfcell{82.86} & \crfcell{32.08}
& \crfcell{78.74} & \crfcell{39.48}
& \crfcell{77.66} & \crfcell{26.48}
& \crfcell{84.21} & \crfcell{44.77}
& \crfcell{83.14} & \crfcell{38.43} \\

& \improw
& \impcell{+11.4} & \impcell{+53.4}
& \impcell{-1.6} & \impcell{+126.2}
& \impcell{+8.2} & \impcell{-14.7}
& \impcell{-2.5} & \impcell{-3.1}
& \impcell{+9.3} & \impcell{+0.4}
& \impcell{+10.8} & \impcell{+25.1}
& \impcell{+5.4} & \impcell{+28.0} \\
\midrule

\multirow{3}{*}{LIQE~\cite{zhang2023blind}}
& w/o PrISM
& 73.58 & 45.55
& 68.20 & 15.80
& 71.27 & 39.34
& 76.77 & 50.79
& 62.39 & 28.31
& 76.21 & 43.51
& 71.90 & 35.32 \\

& \crfcell{w/ PrISM}
& \crfcell{93.84} & \crfcell{53.02}
& \crfcell{90.68} & \crfcell{34.94}
& \crfcell{83.75} & \crfcell{45.87}
& \crfcell{87.57} & \crfcell{51.96}
& \crfcell{77.68} & \crfcell{31.58}
& \crfcell{87.73} & \crfcell{50.85}
& \crfcell{87.74} & \crfcell{44.59} \\

& \improw
& \impcell{+27.5} & \impcell{+16.4}
& \impcell{+33.0} & \impcell{+121.1}
& \impcell{+17.5} & \impcell{+16.6}
& \impcell{+14.1} & \impcell{+2.3}
& \impcell{+24.5} & \impcell{+11.6}
& \impcell{+15.1} & \impcell{+16.9}
& \impcell{+22.0} & \impcell{+26.2} \\
\midrule

\multirow{3}{*}{TOPIQ~\cite{chen2024topiq}}
& w/o PrISM
& 78.38 & 53.46
& 74.29 & 32.92
& 73.50 & {46.77}
& 73.46 & 50.96
& 62.52 & 26.88
& 77.11 & 43.96
& 70.53 & 33.92 \\

& \crfcell{w/ PrISM}
& \crfcell{90.82} & \crfcell{43.27}
& \crfcell{81.90} & \crfcell{16.35}
& \crfcell{83.78} & \crfcell{37.32}
& \crfcell{83.06} & \crfcell{44.04}
& \crfcell{76.99} & \crfcell{28.06}
& \crfcell{85.82} & \crfcell{44.35}
& \crfcell{83.97} & \crfcell{34.30} \\

& \improw
& \impcell{+15.9} & \impcell{-19.1}
& \impcell{+10.2} & \impcell{-50.3}
& \impcell{+14.0} & \impcell{-20.2}
& \impcell{+13.1} & \impcell{-13.6}
& \impcell{+23.1} & \impcell{+4.4}
& \impcell{+11.3} & \impcell{+0.9}
& \impcell{+19.1} & \impcell{+1.1} \\

\midrule

\multirow{3}{*}{Q-Insight~\cite{li2025qinsight}}
& w/o PrISM
& 75.77 & 49.79
& 74.48 & 18.19
& 73.47 & 39.70
& 73.10 & 44.35
& 63.72 & 29.98
& 76.27 & 41.30
& 73.84 & 35.30
\\
& \crfcell{w/ PrISM}
& \crfcell{90.81} & \crfcell{51.76}
& \crfcell{85.68} & \crfcell{17.54}
& \crfcell{85.60} & \crfcell{40.93}
& \crfcell{82.21} & \crfcell{43.16}
& \crfcell{76.27} & \crfcell{29.95}
& \crfcell{85.26} & \crfcell{43.95}
& \crfcell{84.92} & \crfcell{36.10}
\\
& \improw
& \impcell{+19.9} & \impcell{+4.0}
& \impcell{+15.0} & \impcell{-3.6}
& \impcell{+16.5} & \impcell{+3.1}
& \impcell{+12.5} & \impcell{-2.7}
& \impcell{+19.7} & \impcell{-0.1}
& \impcell{+11.8} & \impcell{+6.4}
& \impcell{+15.0} & \impcell{+2.3}
\\
\bottomrule
\end{tabular}}
\end{table}

\noindent\textbf{Evaluation Metrics.}
We report accuracy (ACC), threshold area under the receiver operating characteristic curve (tAUC), and quadratic weighted kappa (QWK) to evaluate overall classification correctness, threshold-based diagnostic accuracy, and ordinal severity agreement, respectively. tAUC is computed as the average AUC over three ordered decision boundaries, while QWK penalizes larger severity disagreements more heavily. Formal definitions of these metrics are given in Appendix~\ref{app:metrics}.

\subsection{Main Results}
\label{sec:main_results}
\noindent\textbf{Results on Reconstructed SPAQ.}
Table~\ref{tab:local_feature_merged} reports the issue-wise performance of PrISM-IQA on reconstructed SPAQ. Two observations are particularly notable. First, different semantic groups have markedly different difficulty. \texttt{Face} is the easiest category in terms of ACC and tAUC, suggesting that semantically salient and visually structured regions are well captured by the vision-language unary scorer. However, its QWK is noticeably lower than its ACC, indicating that while the model reliably detects the presence of face-related defects, the boundary between \texttt{minor} and \texttt{severe} remains more subtle. Second, \texttt{dominant\_color} is the most challenging category, with the lowest tAUC and QWK. This is expected because dominant-color degradations are weakly localized, highly scene-dependent, and often entangled with illumination and material appearance. In other words, the same chromatic deviation may be perceived differently depending on scene semantics, making ordinal annotation and prediction intrinsically harder.

At the issue level, labels such as \texttt{brightness\_high}, \texttt{brightness\_low}, and \texttt{highlight\_clipped} are comparatively easier, likely because they correspond to visually distinct ISP failures with relatively stable perceptual signatures. By contrast, subtle contrast and color-rendering issues remain more difficult. For instance, \texttt{contrast\_low} and some saturation-related labels have substantially lower QWK, which suggests that the model can often recognize the relevant failure mode but has greater difficulty calibrating its exact severity. Overall, these results support the intended use case of PrISM-IQA: the model is strongest on operationally salient failures and still remains reasonably stable across a heterogeneous set of global and local issues. In addition to these diagnostic results, Appendix~\ref{app:openisp_tuning} presents an ISP-tuning case study, showing how PrISM-IQA predictions can be translated into actionable OpenISP adjustments.

\noindent\textbf{Results on the Expert-Annotated Dataset.} 
Table~\ref{tab:expert_results} reports the results on the expert-annotated dataset. Compared with reconstructed SPAQ, this dataset has narrower issue coverage but stronger label reliability, making it particularly useful for evaluating whether the proposed formulation transfers to expert-defined operational defects. Overall, PrISM-IQA maintains strong performance on this independently curated dataset, indicating that its predictions remain reliable for threshold-based diagnosis and ordinal severity estimation under real smartphone capture conditions.

A noteworthy pattern is that these  results are broadly consistent with the trends observed on reconstructed SPAQ in Table~\ref{tab:local_feature_merged}. In particular, PrISM-IQA remains strong on white-balance labels, such as \texttt{wb\_red\_cast}, suggesting that chromatic-direction defects are learned in a way that transfers from the reconstructed setting to real expert-labeled captures. At the same time, labels related to scalar tone changes, such as \texttt{brightness\_low}, remain relatively more challenging. This indicates that the transfer behavior is not simply explained by the presence of exclusion constraints, since both white-balance and tone-related issues involve structured label relations. Instead, the results suggest that PrISM-IQA generalizes better when the visual manifestation of an issue has a more distinctive directional cue, while issues whose severity depends on global calibration or subjective tolerance remain harder across both synthetic and real settings.

\subsection{Ablation Studies}
\noindent\textbf{Generality across NR-IQA Models.}
Rather than treating existing NR-IQA models only as competing baselines, we use them to test whether the proposed structured multi-issue ordinal inference can consistently improve issue-level diagnosis across diverse unary scorers, as summarized in Table~\ref{tab:main_comparison}. The evaluated models cover a broad range of design choices, including two-stream convolutional representations~\cite{zhang2018blind}, multi-scale Transformer encoding~\cite{ke2021musiq}, convolution-Transformer hybrids~\cite{golestaneh2022no}, uncertainty-aware pairwise quality ranking~\cite{zhang2021uncertainty}, vision-language correspondence~\cite{zhang2023blind}, semantic-distortion region modeling~\cite{chen2024topiq}, and reasoning-induced quality assessment~\cite{li2025qinsight}. 

A consistent pattern across these architectures is that PrISM-based structured inference improves tAUC and often QWK. The repeated improvement across heterogeneous NR-IQA backbones indicates that the benefit of the structured inference layer is not tied to a specific feature extractor or scoring architecture, but instead comes from enforcing issue-level ordinal and semantic consistency. Additional baseline comparisons on the expert-annotated dataset are provided in Appendix~\ref{app:aread}.

\begin{wraptable}{r}{0.30\textwidth}
\centering
\caption{Linear probing for MOS prediction on SPAQ. All feature extractors are frozen and paired with the same linear regressor. Best SRCC/PLCC values are bold.}
\label{tab:linear_probe_spaq}
\scriptsize
\begin{tabular}{lcc}
\toprule
Method & SRCC$\uparrow$ & PLCC$\uparrow$ \\
\midrule
\rowcolor{summarybg}
\multicolumn{3}{l}{\textit{Handcrafted baselines}} \\
NIQE~\cite{mittal2013niqe} & 0.5051 & 0.5068 \\
\rowcolor{summarybg}
\multicolumn{3}{l}{\textit{Pretrained visual backbones}} \\
DINOv2~\cite{oquab2023dinov2} & 0.7390 & 0.7464 \\
MAE~\cite{he2022masked} & 0.8453 & 0.8514 \\
iBOT~\cite{zhou2021ibot} & 0.8417 & 0.8471 \\
\rowcolor{summarybg}
\multicolumn{3}{l}{\textit{Pretrained vision-language baselines}} \\
CLIP~\cite{radford2021learning} & 0.8345 & 0.8408 \\
\hline
PrISM-IQA & \textbf{0.8819} & \textbf{0.8873} \\
\bottomrule
\end{tabular}
\end{wraptable}

\noindent\textbf{Perceptual Relevance via Linear Probing.}
We additionally examine whether the representation learned by PrISM-IQA preserves broader perceptual quality information beyond the proposed issue-level diagnosis task. To this end, we conduct a linear probing experiment for MOS prediction on original SPAQ. For all learnable models, the feature extractor is frozen and the same linear regressor is trained to predict MOS, ensuring that the comparison reflects representation quality rather than differences in regression capacity.

As shown in Table~\ref{tab:linear_probe_spaq}, PrISM-IQA achieves the best performance, with a Spearman's rank correlation coefficient (SRCC)
of $0.8819$ and a Pearson linear correlation coefficient (PLCC) of $0.8873$. It clearly outperforms handcrafted NR-IQA baseline NIQE~\cite{mittal2013niqe}, as well as strong pretrained visual backbones including DINOv2~\cite{oquab2023dinov2}, MAE~\cite{he2022masked}, and iBOT~\cite{zhou2021ibot}. PrISM-IQA also improves over its pretrained vision-language representations from CLIP~\cite{radford2021learning}.

This result is remarkable because the probing task is not the direct training objective of PrISM-IQA: the model is trained using issue-level ordinal supervision rather than MOS regression. The strong linear probing performance suggests that the proposed supervision does not merely encode discrete defect labels. Instead, it encourages the encoder to organize images along perceptually meaningful axes that remain useful for holistic quality prediction. The gains over both generic visual backbones and generic vision-language models further indicate that issue-aware ordinal learning provides complementary perceptual structure beyond standard pretraining.

\section{Conclusion and Discussion}
We have presented PrISM-IQA, an ISP-aware formulation of smartphone IQA that moves beyond conventional scalar quality prediction toward structured, issue-level diagnosis. Instead of assigning an image a single MOS-like score, PrISM-IQA predicts an ordinal severity level for each ISP-relevant quality issue, including both global image-level artifacts and local content-dependent defects. This design better matches practical ISP tuning workflows, where engineers need to identify the specific failure mode, estimate its severity, and determine whether corrective action is required. To make these predictions logically consistent, we combine cumulative ordinal encoding with HEX-constrained structured inference, thereby enforcing within-issue monotonicity as well as cross-issue relations such as subsumption and mutual exclusion. Instantiated with a vision-language unary scorer, the resulting method provides interpretable, ordinal, and semantically meaningful quality diagnosis.

Experiments on reconstructed SPAQ and expert-annotated data demonstrate the effectiveness of PrISM-IQA for fine-grained issue-level assessment. PrISM-IQA performs strongly across diverse global and local issues, transfers to expert-labeled real-world captures, and benefits from structured inference across multiple NR-IQA backbones. Moreover, linear probing for MOS prediction indicates that our formulation does not merely fit discrete defect labels; rather, it encourages the model to learn perceptually meaningful representations that remain useful for holistic quality prediction. Beyond diagnosis, we further show that these structured predictions can support actionable ISP tuning, with the resulting tuned images preferred in most comparisons. These findings suggest that structured diagnostic supervision can provide a practical and informative alternative to scalar IQA objectives, especially in settings where model outputs must support downstream engineering decisions.

Looking ahead, PrISM-IQA can be developed from a defect-diagnosis model into a principled optimization framework for smartphone photography. A first direction is to expand the issue taxonomy beyond low-level ISP artifacts toward higher-level photographic intentions and perceptual styles, such as atmosphere, cinematic rendering, documentary realism, naturalness, subject emphasis, and emotional tone. These abstract attributes are difficult to define with fixed scalar labels, making them well suited to structured ordinal or preference-based supervision collected from expert raters, pairwise comparisons, and human-in-the-loop active learning. A second direction is to close the loop between quality assessment and ISP tuning in an end-to-end manner. Rather than using manually designed issue-to-module rules, future systems should learn differentiable or surrogate ISP controllers that translate predicted issue severities and aesthetic targets into parameter-level updates for exposure, tone mapping, white balance, denoising, sharpening, color rendering, and local enhancement. This would allow IQA predictions to become optimization objectives, enabling constrained multi-objective tuning that balances defect removal, aesthetic intent, device consistency, and scene-dependent user preference. A third direction is to make this optimization reliable in real deployment by incorporating uncertainty estimation, counterfactual diagnostics, temporal consistency for video, cross-device adaptation, and expert-validated benchmarks that jointly evaluate diagnostic accuracy, ISP actionability, and human preference. These directions would move PrISM-IQA from structured assessment toward closed-loop, interpretable, and preference-aligned computational photography, where quality models not only identify what is wrong with an image but also guide how a camera pipeline should improve it.

{
    \small
    \bibliographystyle{plainnat}
    \bibliography{refs}
}


\newpage

\appendix

\section*{\Large\bfseries Appendix}

This appendix supplements the main paper with practical demonstrations, reproducibility details, and additional empirical comparisons. We first show how PrISM-IQA can be used beyond diagnosis by translating its structured issue-level predictions into OpenISP tuning actions, followed by model-based re-evaluation and a small subjective study. We then provide the experimental setups in detail, including dataset construction, compute resources, and the formal definitions of the evaluation metrics. Finally, we report additional baseline comparisons on the expert-annotated dataset to further assess PrISM-IQA under real smartphone capture conditions.

\section{PrISM-IQA-Guided ISP Tuning with OpenISP}
\label{app:openisp_tuning}

In this section, we examine whether the structured issue-level predictions of PrISM-IQA can provide actionable feedback for ISP tuning. We instantiate this idea with OpenISP, a controllable ISP testbed, by mapping predicted issue severities to interpretable module-level updates.

\noindent\textbf{OpenISP testbed and protocol.}
OpenISP provides a modular simulation of the ISP pipeline, making it suitable for controlled tuning studies. Our experiment follows an IQA-ISP-IQA loop. Given an input image \(x\), PrISM-IQA first predicts the issue-wise severity vector \( f(x) = [\hat y_1,\ldots,\hat y_N]\). We select active issues with \(\hat y_j > 1\), map them to OpenISP tuning actions using a fixed issue-to-module rule table, and scale the update magnitude by the predicted ordinal level. Minor issues induce conservative adjustments, whereas severe and critical issues induce larger changes. OpenISP then renders the tuned image, which PrISM-IQA re-evaluates to check whether target severities decrease and whether new salient regressions appear.

\begin{figure}[t]
\centering
\includegraphics[width=\linewidth]{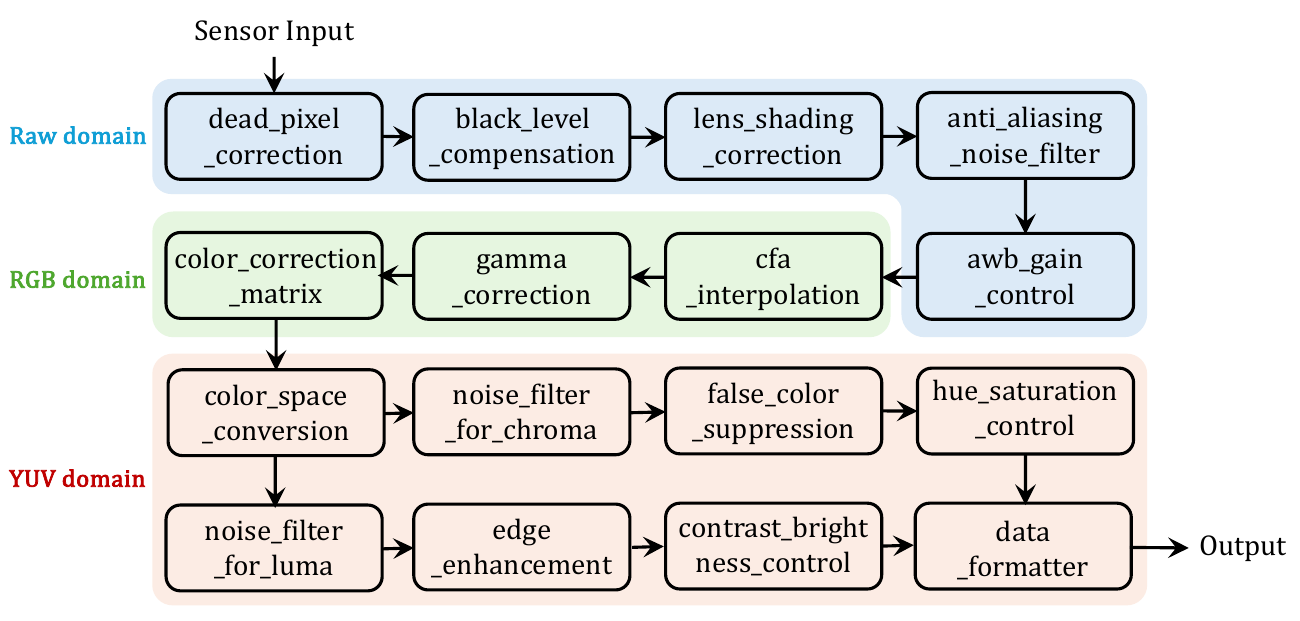}
\caption{
OpenISP processing pipeline used for PrISM-IQA-guided ISP tuning. The pipeline is organized into Raw, RGB, and YUV domains, progressing from sensor-level correction and preprocessing, through demosaicing and color rendering, to luma/chroma refinement, edge enhancement, tone adjustment, and output formatting. This modular design allows for PrISM-IQA's issue-level quality predictions to be translated into interpretable, module-specific ISP adjustments.
}
\label{fig:isp_pipeline}
\end{figure}

\noindent\textbf{Issue-to-module correspondence.}\providecommand{\code}[1]{{\urlstyle{tt}\nolinkurl{#1}}} As shown in Fig.~\ref{fig:isp_pipeline}, we use OpenISP modules with direct perceptual counterparts in the PrISM-IQA taxonomy: 1) \code{dead_pixel_correction}, 2) \code{black_level_compensation}, 3) \code{lens_shading_correction}, 4) \code{anti_aliasing_noise_filter}, 5) \code{awb_gain_control}, 6) \code{cfa_interpolation}, 7) \code{gamma_correction}, 8) \code{color_correction_matrix}, 9) \code{noise_filter_for_luma},  10) \code{edge_enhancement},  11) \code{contrast_brightness_control}, 12) \code{noise_filter_for_chroma}, 13) \code{false_color_suppression}, and 14) \code{hue_saturation_control}. Modules such as \code{color_space_conversion} are kept fixed, since they mainly change representation rather than directly addressing the perceptual failures.

The mapping follows the semantics of the issue taxonomy. Exposure and contrast failures map to \texttt{gamma\_correction} and \texttt{contrast\_brightness\_control}, with \texttt{black\_level\_compensation} and \texttt{lens\_shading\_correction} additionally invoked when low-level illumination non-uniformity is suspected. Highlight failures map to \texttt{gamma\_correction} and \texttt{contrast\_brightness\_control}, sometimes coupled with \texttt{awb\_gain\_control} or \texttt{color\_correction\_matrix} to stabilize highlight color. White-balance and color-cast issues map to \texttt{awb\_gain\_control} and \texttt{color\_correction\_matrix}, optionally with \texttt{lens\_shading\_correction} and \texttt{hue\_saturation\_control} for color non-uniformity or local tint. Saturation issues map to \texttt{hue\_saturation\_control} together with \texttt{color\_correction\_matrix}. Detail and clarity losses map to \texttt{cfa\_interpolation}, \texttt{noise\_filter\_for\_luma}, and \texttt{edge\_enhancement}; excessive sharpness and sharpening artifacts map mainly to \texttt{cfa\_interpolation} and \texttt{edge\_enhancement}. Noise-related issues map to \texttt{dead\_pixel\_correction}, \texttt{anti\_aliasing\_noise\_filter}, \texttt{noise\_filter\_for\_luma}, and  \texttt{noise\_filter\_for\_chroma}. Table~\ref{tab:openisp_issue_module_map} gives the full rule table.

\begin{figure}[t]
\centering
\begin{subfigure}{0.23\textwidth}
        \centering
        \includegraphics[width=\textwidth]{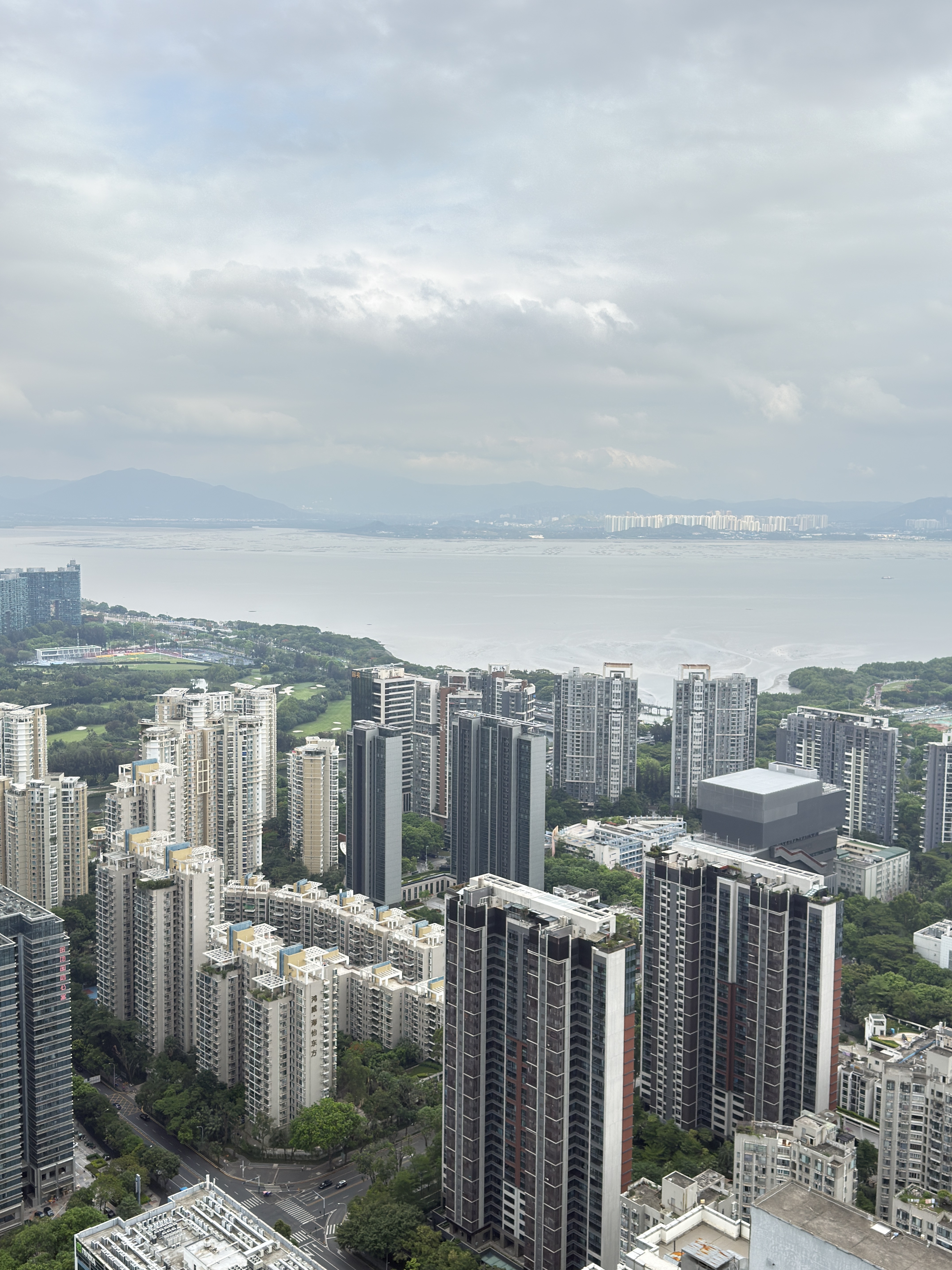}
        \caption{}
        \label{fig:a}
    \end{subfigure}
    \hfill
    \begin{subfigure}{0.23\textwidth}
        \centering
        \includegraphics[width=\textwidth]{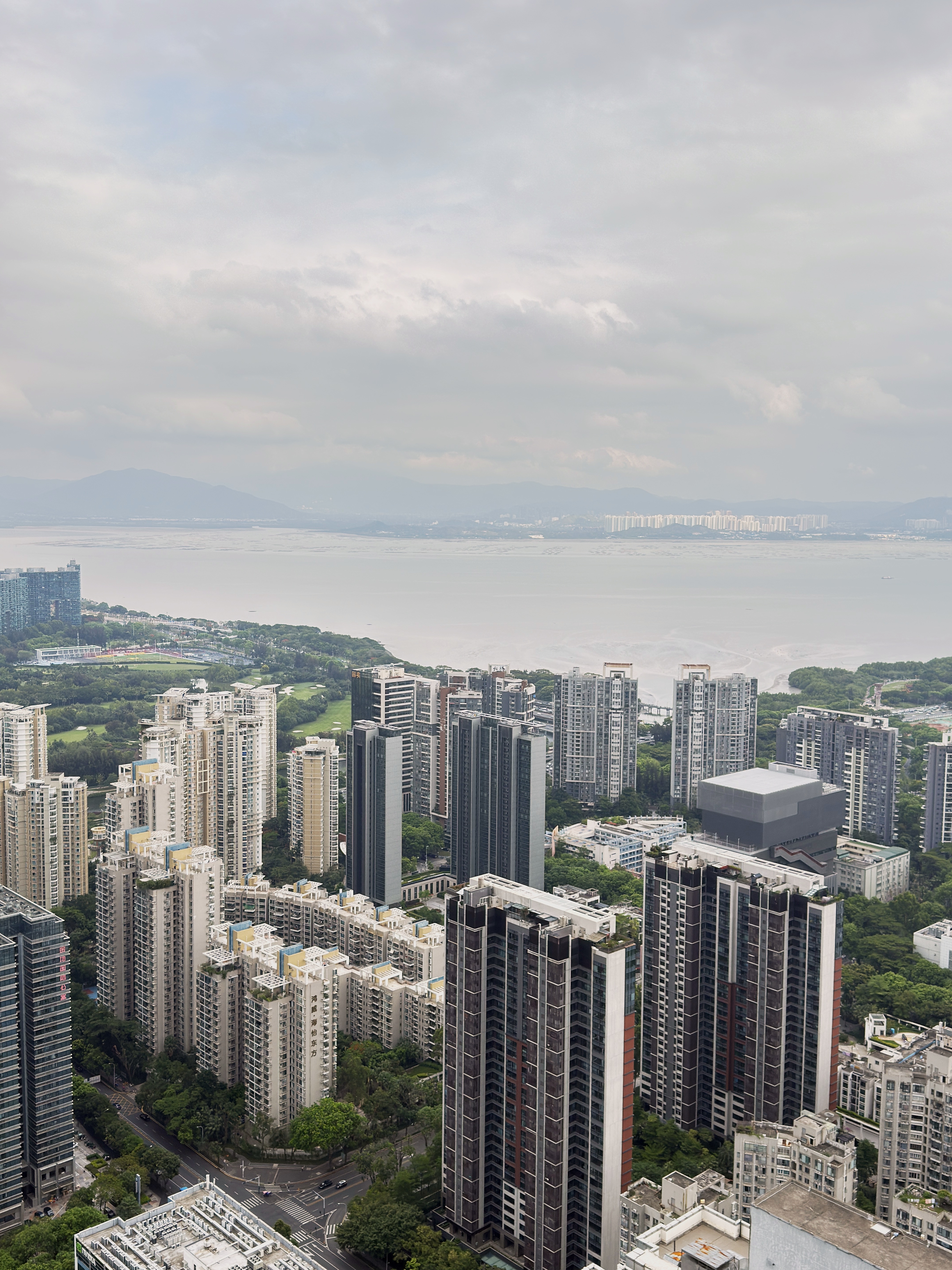}
        \caption{}
        \label{fig:b}
    \end{subfigure}
    \hfill
    \begin{subfigure}{0.23\textwidth}
        \centering
        \includegraphics[width=\textwidth]{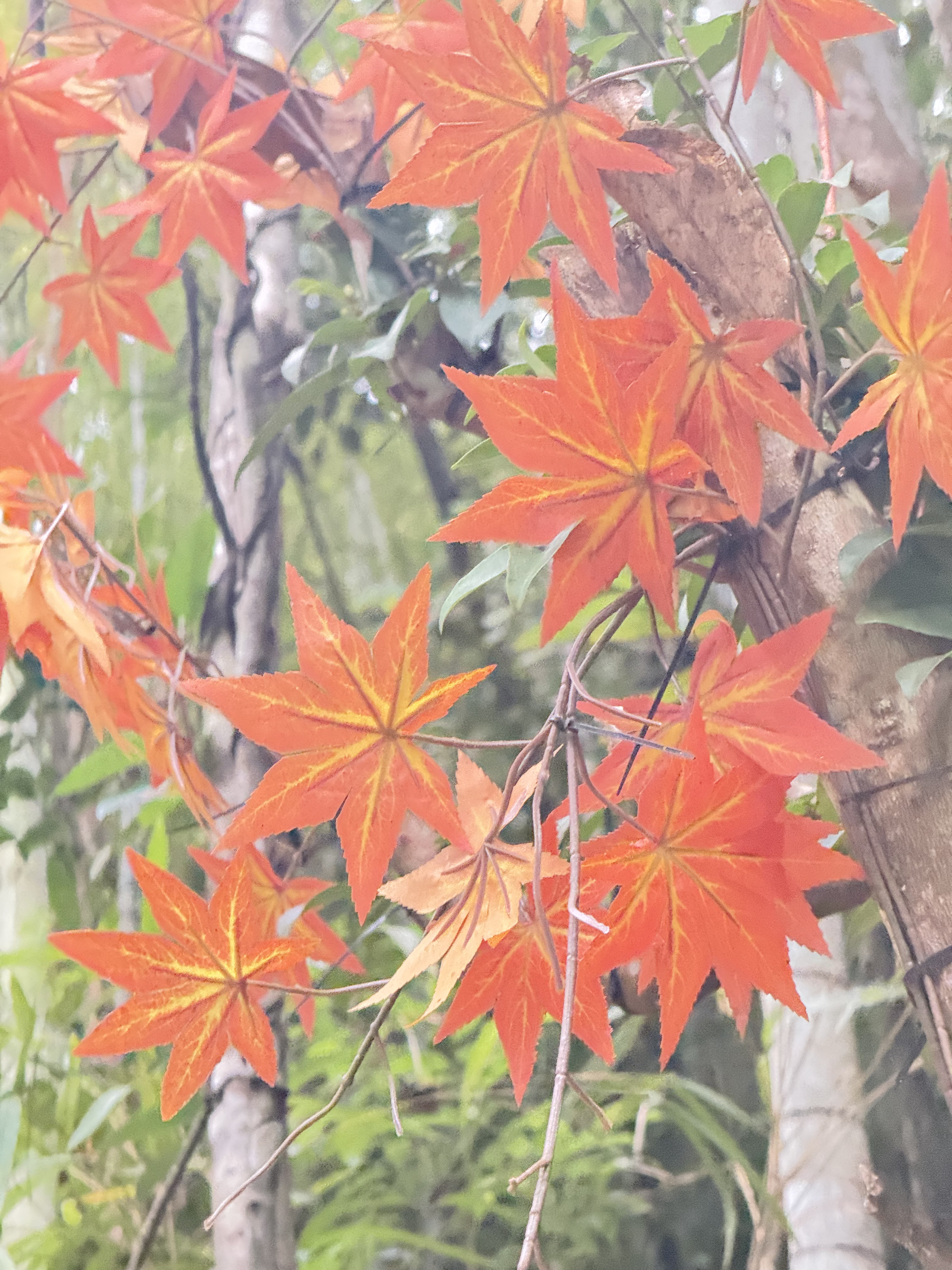}
        \caption{}
        \label{fig:c}
    \end{subfigure}
    \hfill
    \begin{subfigure}{0.23\textwidth}
        \centering
        \includegraphics[width=\textwidth]{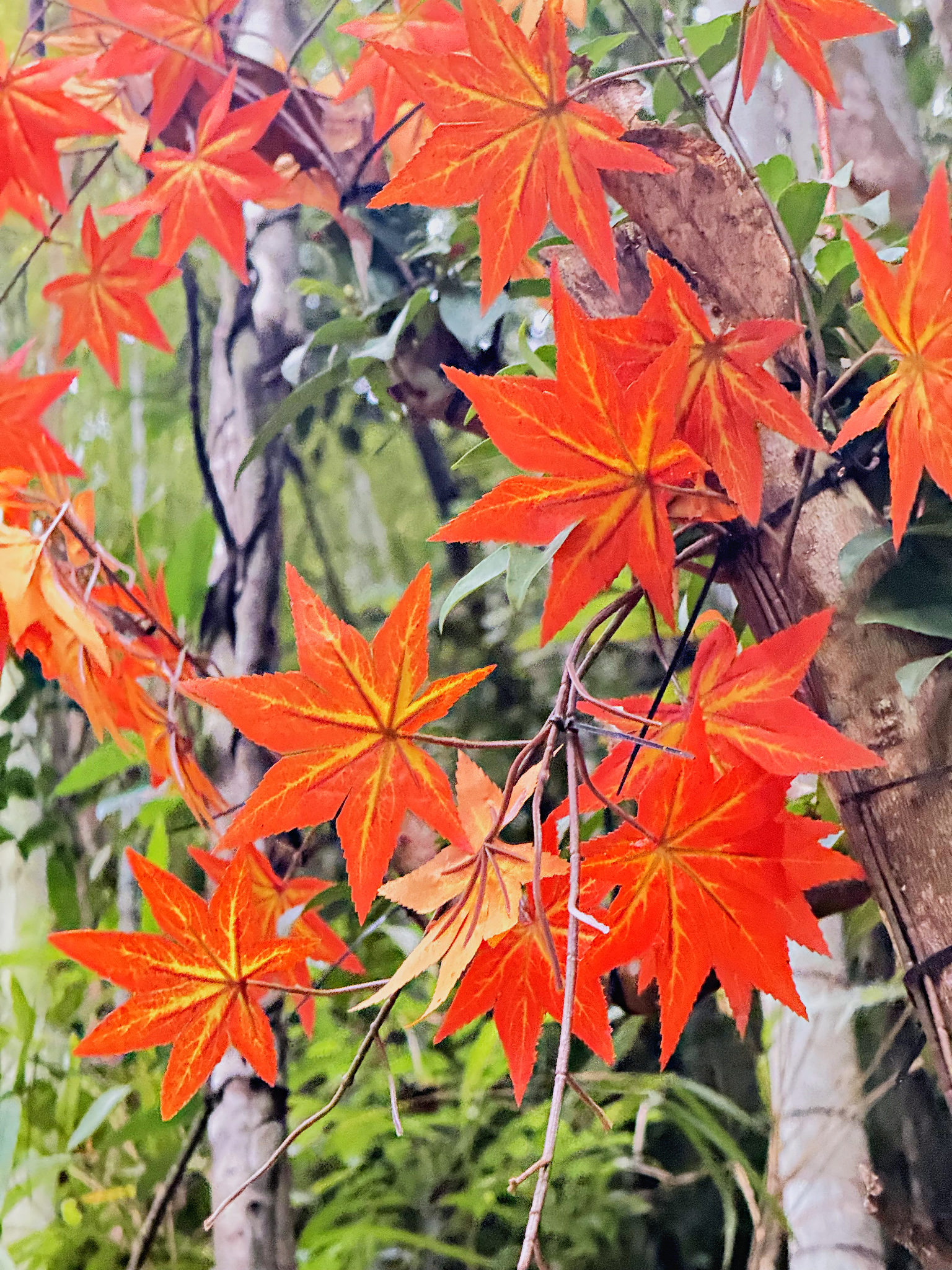}
        \caption{}
        \label{fig:d}
    \end{subfigure}
\caption{Representative examples of PrISM-IQA-guided OpenISP tuning. Panels (a) and (c) show the initial images before tuning, and panels (b) and (d) show the corresponding results after tuning. For Example 1, PrISM-IQA re-evaluation indicates that \texttt{wb\_green\_cast} is reduced from \textit{severe} to \textit{minor} and \texttt{sky\_contrast\_low} from \textit{critical} to \textit{minor}. For Example 2, \texttt{clarity\_low} and \texttt{greenery\_clarity\_low} are reduced from \textit{severe} to \textit{minor} as well, while \texttt{greenery\_contrast\_low}, \texttt{brightness\_high}, and \texttt{contrast\_low}  are eliminated.}
\label{fig:openisp_cases}
\end{figure}

\noindent\textbf{Case studies.}
Fig.~\ref{fig:openisp_cases} shows two examples before/after tuning. In both cases, we diagnose the original image using PrISM-IQA, convert the active issues to OpenISP module updates using Table~\ref{tab:openisp_issue_module_map}, and re-evaluate the tuned result by the same model.

\textbf{Example 1.}
PrISM-IQA identifies \texttt{wb\_green\_cast} as \emph{severe} and \texttt{sky\_contrast\_low} as \emph{critical}. The first issue activates \code{awb_gain_control}, with \code{color_correction_matrix} and \code{hue_saturation_control} as auxiliary modules. The second activates \code{contrast_brightness_control}, with \code{gamma_correction} as an auxiliary module. The resulting OpenISP update combines mild white-balance correction, a small shift toward neutral color, and a tone adjustment that increases sky contrast. Concretely, it slightly increases the red gain, reduces the green and blue gains, applies a mild color correction, decreases brightness, compresses highlights, increases global and sky contrast, and adds conservative saturation and sharpness adjustments. After tuning, PrISM-IQA re-labels both \texttt{wb\_green\_cast} and \texttt{sky\_contrast\_low} as \emph{minor}. The tuned image also better matches the scene appearance: the overcast sky becomes darker and less blue-green, and building details are clearer.

\textbf{Example 2.}
The second image contains a broader mixture of exposure, contrast, and detail failures. PrISM-IQA predicts \texttt{greenery\_contrast\_low} at the \emph{critical} level, and \texttt{greenery\_clarity\_low}, \texttt{brightness\_high}, \texttt{contrast\_low}, and \texttt{clarity\_low} at the \emph{severe} level. The exposure and global contrast issues activate \code{gamma_correction} and \code{contrast_brightness_control}; the local greenery contrast issue activates the same tone modules; and the clarity issues activate \code{cfa_interpolation}, \code{noise_filter_for_luma}, and \code{edge_enhancement}. We also apply mild \code{awb_gain_control}, \code{color_correction_matrix}, and \code{hue_saturation_control} updates to stabilize foliage color while preserving the red leaves.

The OpenISP update reduces the washed, over-bright appearance by decreasing brightness and compressing highlights, restores global and local contrast with mild gamma and contrast increases, and improves fine structure through conservative edge enhancement with weak denoising. 
After tuning, PrISM-IQA predicts that \texttt{clarity\_low} and \texttt{greenery\_clarity\_low} are reduced to \emph{minor}, while \texttt{greenery\_contrast\_low}, \texttt{brightness\_high}, and \texttt{contrast\_low} are removed. This example illustrates how multiple predicted issues can be converted into coordinated ISP updates across exposure, contrast, color, and detail modules.

\noindent\textbf{Subjective validation.}
We further conduct a small paired comparison study to test whether the tuned outputs are perceptually preferred. Five participants---one female and four males, aged 20–40 years---who have general knowledge of image processing and computer vision but are unaware of the study’s objective compare $20$ image pairs before and after tuning, yielding 100 judgments in total. For each pair, the participant chooses whether the OpenISP-tuned result is better, the initial is better, or the perceptual difference is negligible. As shown in Table~\ref{tab:openisp_user}, 
the tuned result receives \(84.0\%\) of all votes and \(93.3\%\) of decisive votes after excluding ties. 

This small study is not intended to replace large-scale ISP optimization. Instead, it provides an end-to-end demonstration that issue-level ordinal IQA can identify actionable defects, map them to interpretable ISP modules, produce tuned outputs that human observers often prefer, and verify the result again using the same issue-level diagnosis model.

\begin{table}[t]
\centering
\footnotesize
\caption{Subjective paired comparison results for PrISM-IQA-guided OpenISP tuning. Five participants evaluate $20$ initial/tuned image pairs, yielding $100$ total judgments. Counts and percentages summarize the distribution of preferences, with the final row reporting the preference rate of tuned images after excluding ties.}
\label{tab:openisp_user}
\begin{tabular}{lcc}
\toprule
Judgment outcome & Count & Percentage \\
\midrule
OpenISP-tuned image preferred & 84 & 84.0\% \\
Initial image preferred & 6 & 6.0\% \\
No perceptible difference & 10 & 10.0\% \\
\midrule
OpenISP-tuned image preferred among decisive judgments & 84 / 90 & 93.3\% \\
\bottomrule
\end{tabular}
\end{table}

\section{Experiment Setups}
This section provides additional implementation details for reproducibility, including dataset construction, compute resource, and metric definition.

\begin{figure}
    \centering
    \includegraphics[width=0.9\linewidth]{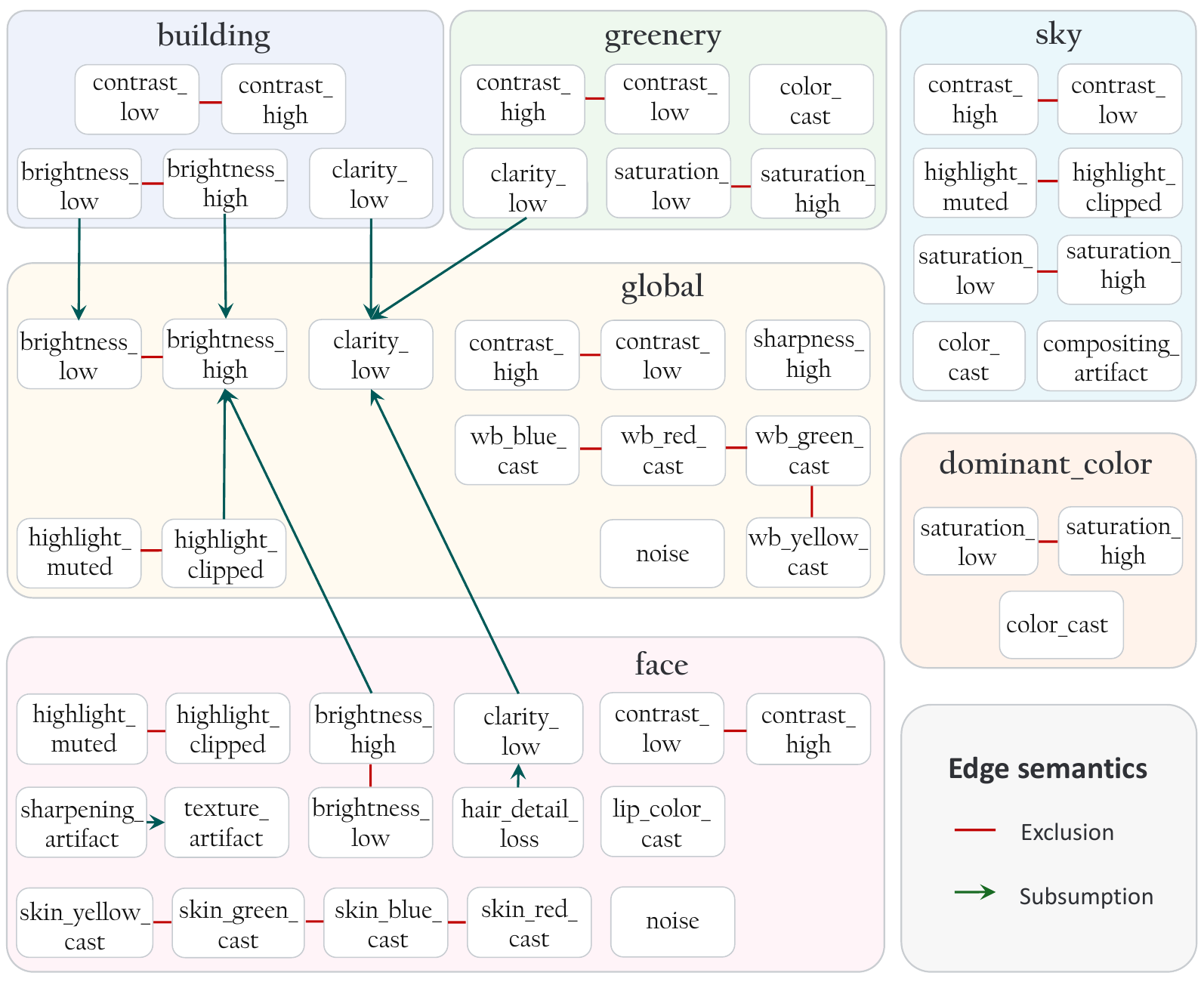}
    \caption{HEX graph for the reconstructed SPAQ issue taxonomy. Colored panels list issue nodes by semantic scope: one global image-level group and five local content-dependent groups. Red links denote mutual exclusion relations, while green arrows summarize subsumption mainly from local issues to their corresponding global parent issues. During inference, each issue node is subjected to cumulative ordinal encoding, where within-issue monotonicity is enforced together with these cross-issue relations.}
    \label{fig:spaq_graph}
\end{figure}

\subsection{Datasets}
\label{app:datasets}
\noindent\textbf{Reconstructed SPAQ.}
We reconstruct SPAQ as a multi-issue ordinal diagnosis benchmark by re-annotating all $11,125$ images with dense labels for $53$ ISP-relevant quality issues. The taxonomy is practical rather than exhaustive: experts with substantial ISP tuning experience select recurrent, actionable failure modes along the smartphone imaging pipeline. The issue space covers exposure and tone-rendering errors (\eg, brightness inaccuracy, contrast imbalance, and highlight mishandling), color and white-balance errors (\eg, cast and saturation problems), and detail-rendering errors (\eg, clarity loss, unnatural texture rendition, and noise-related defects). In addition to global artifacts, it also includes local issues for semantically important regions: \texttt{building}, \texttt{face}, \texttt{greenery}, \texttt{sky}, and \texttt{dominant\_color}. Although the labels are defined by observable failures rather than algorithmic modules, they loosely correspond to ISP stages, as demonstrated in Table~\ref{tab:openisp_issue_module_map}.

For each image, we assign a four-level severity label---\emph{absent}, \emph{minor}, \emph{severe}, or \emph{critical}---to each global image-level or local content-dependent issue. For region-specific labels, annotations are made only when the corresponding semantic region is present. If an image contains no face, for example, face-related labels are treated as \textit{non-applicable} and \textit{undefined} rather than assigned an \emph{absent} severity. The re-annotation is performed with GPT-5, yielding dense labels for all $53$ issues on every image. We also construct the HEX graph in Fig.~\ref{fig:spaq_graph}, which encodes ordinal monotonicity, mutual exclusion, and subsumption relations for structured inference. Among the $53$ issues, $19$ appear in fewer than $100$ images (\ie, less than $1\%$ of the dataset) and show noticeable annotation noise under human inspection. To avoid unstable supervision, we treat these issues as unlabeled during training, and exclude them from quantitative evaluation. The reconstructed dataset is split into training, validation, and test sets with a ratio of $7$:$1$:$2$.

\begin{figure}[t]
    \centering
    \captionsetup[subfigure]{labelformat=parens,labelsep=none}

    \begin{subfigure}[t]{0.19\linewidth}
        \centering
        \includegraphics[width=\linewidth]{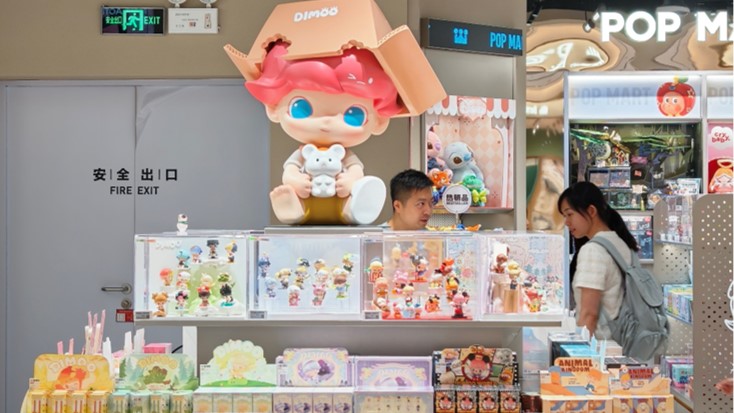}
        \caption{}
    \end{subfigure}\hfill
    \begin{subfigure}[t]{0.19\linewidth}
        \centering
        \includegraphics[width=\linewidth]{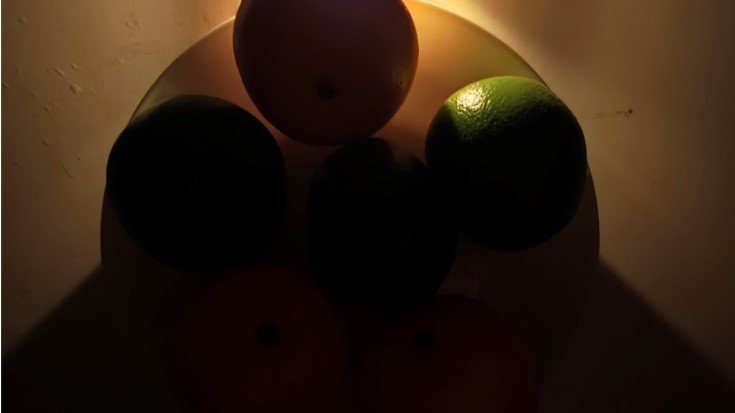}
        \caption{}
    \end{subfigure}\hfill
    \begin{subfigure}[t]{0.19\linewidth}
        \centering
        \includegraphics[width=\linewidth]{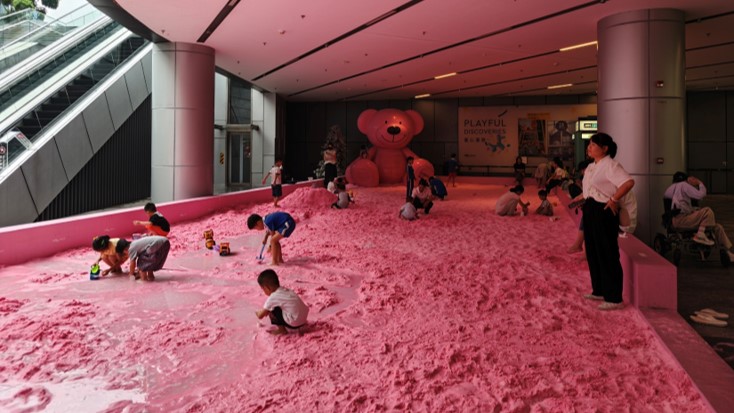}
        \caption{}
    \end{subfigure}\hfill
    \begin{subfigure}[t]{0.19\linewidth}
        \centering
        \includegraphics[width=\linewidth]{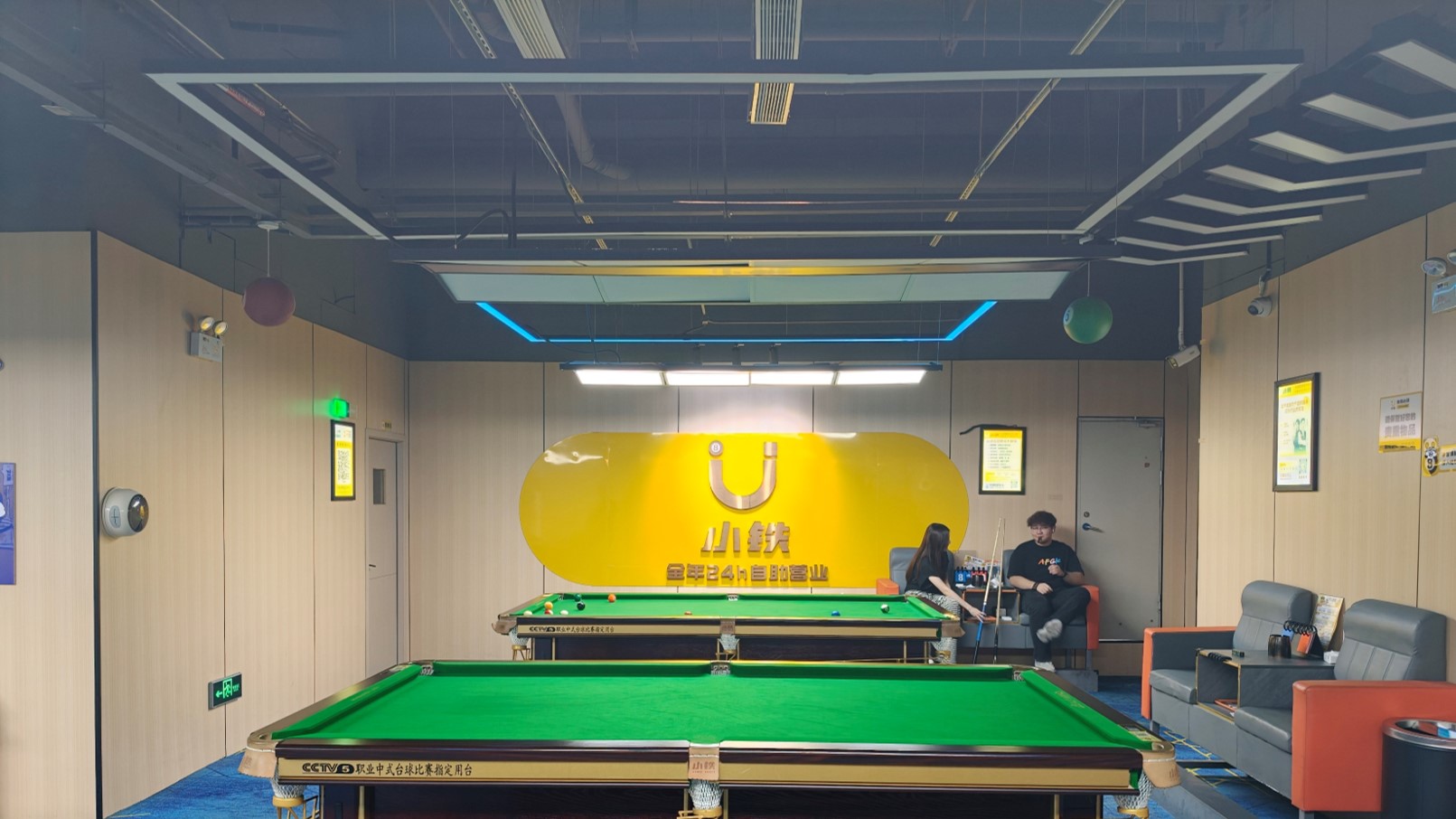}
        \caption{}
    \end{subfigure}\hfill
    \begin{subfigure}[t]{0.19\linewidth}
        \centering
        \includegraphics[width=\linewidth]{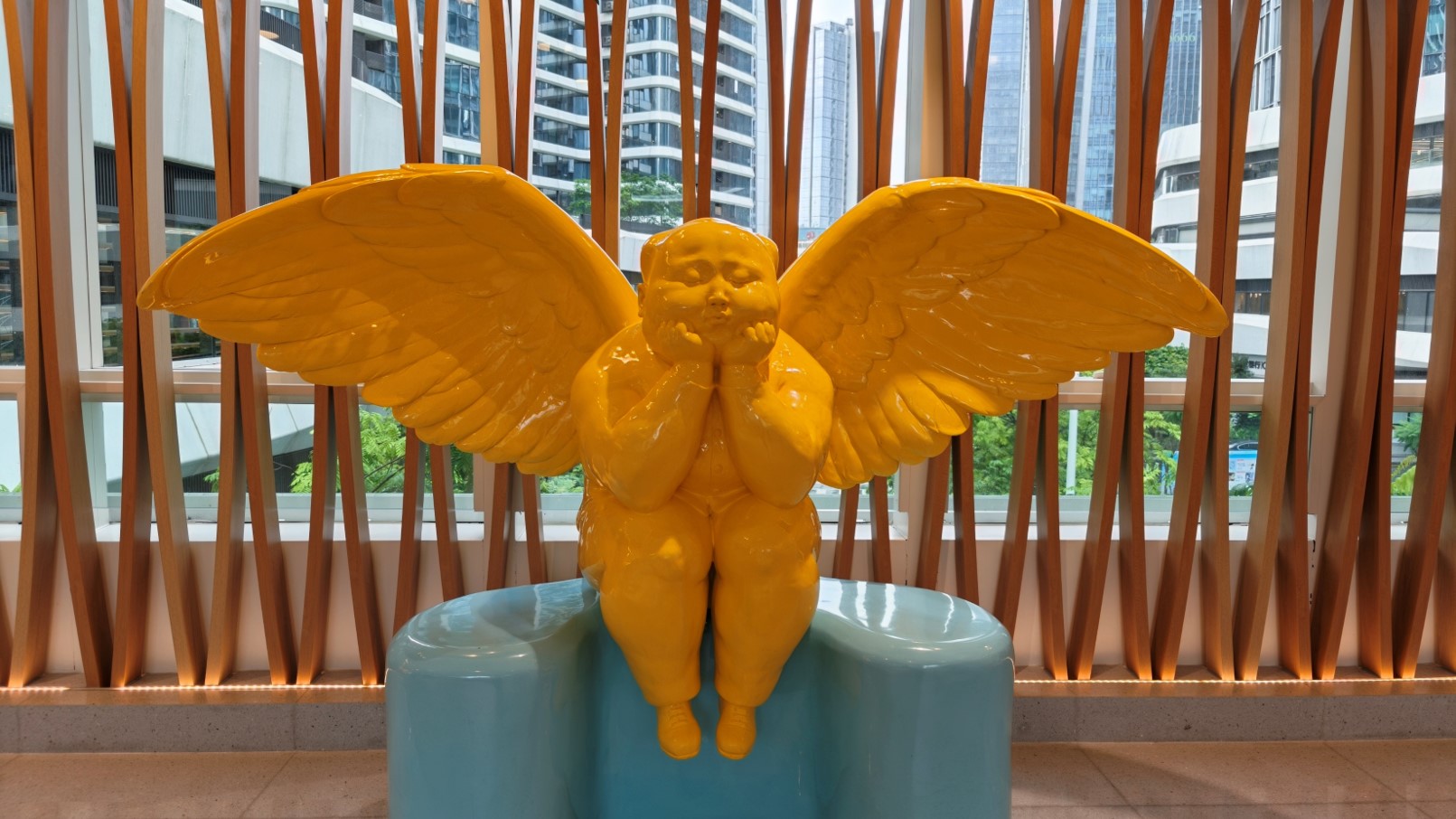}
        \caption{}
    \end{subfigure}

    \vspace{1mm}

    \begin{subfigure}[t]{0.19\linewidth}
        \centering
        \includegraphics[width=\linewidth]{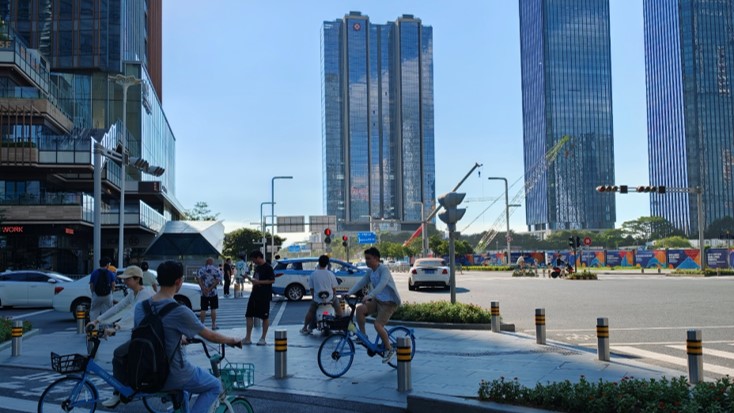}
        \caption{}
    \end{subfigure}\hfill
    \begin{subfigure}[t]{0.19\linewidth}
        \centering
        \includegraphics[width=\linewidth]{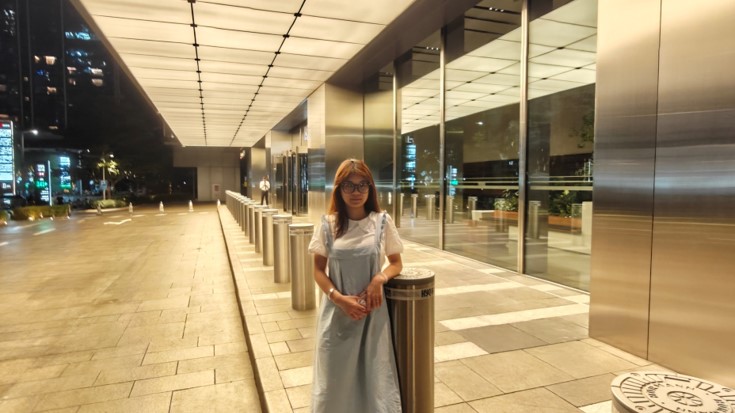}
        \caption{}
    \end{subfigure}\hfill
    \begin{subfigure}[t]{0.19\linewidth}
        \centering
        \includegraphics[width=\linewidth]{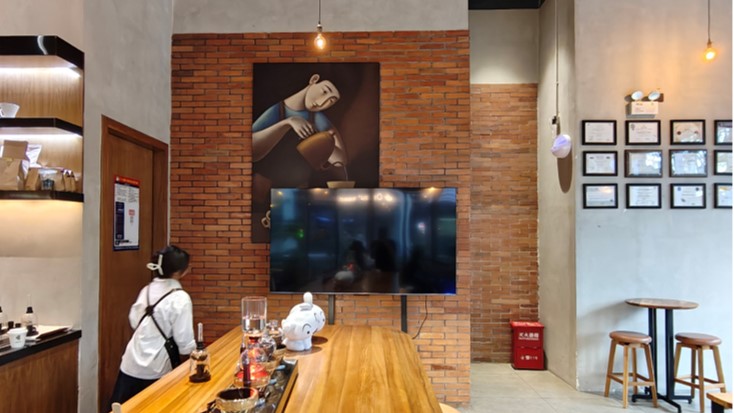}
        \caption{}
    \end{subfigure}\hfill
    \begin{subfigure}[t]{0.19\linewidth}
        \centering
        \includegraphics[width=\linewidth]{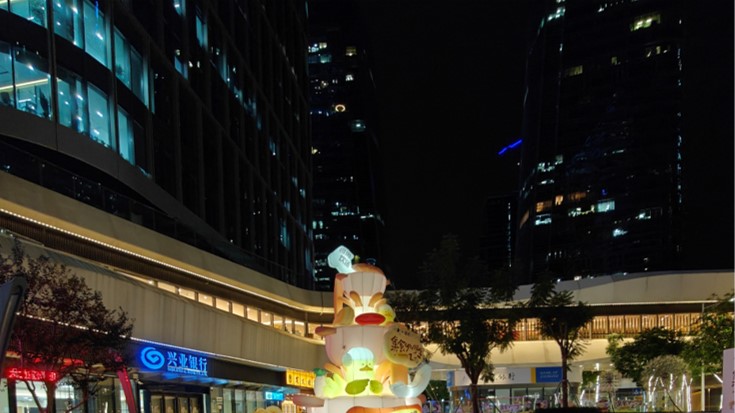}
        \caption{}
    \end{subfigure}\hfill
    \begin{subfigure}[t]{0.19\linewidth}
        \centering
        \includegraphics[width=\linewidth]{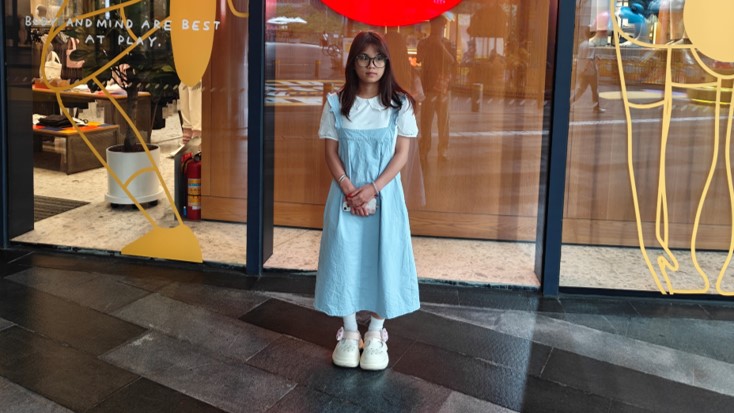}
        \caption{}
    \end{subfigure}

    \caption{Representative images from the expert-annotated dataset. Panels illustrate common global image-level issues, including (a) \texttt{brightness\_high}, (b) \texttt{brightness\_low}, (c) \texttt{contrast\_high}, (d) \texttt{contrast\_low}, (e) \texttt{wb\_blue\_cast}, (f) \texttt{wb\_red\_cast}, and (g) \texttt{wb\_yellow\_cast}. Panels (h)-(j) show increasing severity levels for \texttt{brightness\_low}, from \textit{minor} to \textit{critical}.}
    \label{fig:data samples}
\end{figure}

\noindent\textbf{Expert-Annotated Data.}
We further construct an expert-annotated dataset of $2,983$ images acquired by imaging experts from a major smartphone manufacturer using three smartphone models. The dataset covers indoor and outdoor scenes, daytime and nighttime environments, and portrait and landscape scenarios. Unlike reconstructed SPAQ, which emphasizes breadth of issue coverage, this dataset emphasizes label reliability and utility. We therefore focus on seven global issues that are operationally important in ISP development and consistently judgeable across scenes: \texttt{brightness\_high}, \texttt{brightness\_low}, \texttt{contrast\_high}, \texttt{contrast\_low}, 
\texttt{wb\_blue\_cast},
\texttt{wb\_red\_cast}, and \texttt{wb\_yellow\_cast}. The exclusion relations are defined between \texttt{brightness\_high} and  \texttt{brightness\_low}, between \texttt{contrast\_high} and \texttt{contrast\_low}, and among the three white-balance cast labels. Example images are shown in Fig.~\ref{fig:data samples}.

Each image is independently annotated by two imaging experts using the same four-level severity rubric as reconstructed SPAQ. Disagreements are resolved by adjudicated consensus with reference to the shared annotation guideline. This protocol is especially important near the operational boundary between \emph{minor} and \emph{severe}, where ISP tuning actions are often triggered in practice. The dataset is split into training, validation, and test sets with a ratio of $7$:$1$:$2$.

\subsection{Compute Resources}
All experiments are conducted on a single NVIDIA A100 80 GB GPU. The machine is equipped with two Intel Xeon Platinum 8336C CPUs at 2.30 GHz, providing 128 CPU threads in total, and 503 GiB system memory. Each training run uses one A100 GPU and takes approximately $3$ hours for $50$ epochs. The software environment uses PyTorch 2.8.0 with CUDA 12.8.

\subsection{Evaluation Metrics}
\label{app:metrics}
For a fixed quality issue \(j\), let \(\mathcal{I}_j\) be the index set of test samples annotated for that issue.

\noindent\textbf{Accuracy (ACC).}
For issue \(j\), accuracy is the fraction of samples whose predicted severity exactly matches the ground-truth severity:
\begin{equation}
\mathrm{ACC}_j
=
\frac{1}{\vert\mathcal{I}_j\vert}
\sum_{i \in \mathcal{I}_j}
\mathbb{I}\!\left[\hat y_j^{(i)} = y_j^{(i)}\right],
\end{equation}
where \(\hat y_j^{(i)}\) is obtained by maximum a posteriori decoding in Eq.~\eqref{eq:map}.

\noindent\textbf{Threshold AUC (tAUC).}
To extend binary AUC to ordinal labels, we evaluate the cumulative thresholds defined in Sec.~\ref{subsec:cumenc}. For issue \(j\) and threshold \(k \in \{1,\ldots,K_j-1\}\), positives are annotated samples with \(b_{j,k}^{(i)}=1\) and negatives are annotated samples with \(b_{j,k}^{(i)}=0\):
\[
\mathcal{I}_{j,k}^{+} = \left\{\, i \in \mathcal{I}_j : b_{j,k}^{(i)} = 1 \,\right\},
\qquad
\mathcal{I}_{j,k}^{-} = \left\{\, i \in \mathcal{I}_j : b_{j,k}^{(i)} = 0 \,\right\}.
\]
Let \(\mathcal{T}_j\) be the set of threshold indices for which both sets are nonempty. Thresholds outside \(\mathcal{T}_j\) are undefined for AUC and are skipped. Using the predicted probability \(P\left(b_{j,k}^{(i)}=1 \mid x^{(i)}\right)\) as the ranking score, the threshold-specific AUC is defined as
\begin{equation}
\begin{aligned}
\mathrm{AUC}_{j,k}
&=
\frac{1}{|\mathcal{I}_{j,k}^{+}|\,|\mathcal{I}_{j,k}^{-}|}
\sum_{i \in \mathcal{I}_{j,k}^{+}}
\sum_{i' \in \mathcal{I}_{j,k}^{-}}
\Bigg(
\mathbb{I}\!\left[
P\left(b_{j,k}^{(i)}=1 \mid x^{(i)}\right)
>
P\left(b_{j,k}^{(i')}=1 \mid x^{(i')}\right)
\right] \\
&\qquad\qquad\qquad\qquad
+
\frac{1}{2}\mathbb{I}\!\left[
P\left(b_{j,k}^{(i)}=1 \mid x^{(i)}\right)
=
P\left(b_{j,k}^{(i')}=1 \mid x^{(i')}\right)
\right]
\Bigg).
\end{aligned}
\end{equation}

The tAUC for issue \(j\) averages over the defined ordinal thresholds:
\begin{equation}
\label{eq:app_tauc}
\mathrm{tAUC}_j
=
\frac{1}{|\mathcal{T}_j|}
\sum_{k \in \mathcal{T}_j} \mathrm{AUC}_{j,k}.
\end{equation}
In the four-level setting used in this paper, the three cumulative events are \(b_{j,1}=1\), \(b_{j,2}=1\), and \(b_{j,3}=1\), corresponding to \textit{mild}, \textit{severe}, and \textit{critical} issues, respectively. 

\begin{table}[t]
\footnotesize
\centering
\caption{Additional comparison on the expert-annotated dataset across seven issue labels. Best values are in \textbf{bold}, and second-best values are \underline{underlined}.}
\label{tab:expert_comparisons}
\small
\renewcommand{\arraystretch}{1.08}
\begin{tabular}{l|cc|cc|cc|cc}
\toprule
\multirow{2}{*}{Method}
& \multicolumn{2}{c|}{\texttt{contrast\_high}}
& \multicolumn{2}{c|}{\texttt{contrast\_low}}
& \multicolumn{2}{c|}{\texttt{brightness\_high}}
& \multicolumn{2}{c}{\texttt{brightness\_low}} \\
\cmidrule(lr){2-3}\cmidrule(lr){4-5}\cmidrule(lr){6-7}\cmidrule(lr){8-9}
& tAUC$\uparrow$ & QWK$\uparrow$
& tAUC$\uparrow$ & QWK$\uparrow$
& tAUC$\uparrow$ & QWK$\uparrow$
& tAUC$\uparrow$ & QWK$\uparrow$ \\
\midrule

MUSIQ~\cite{ke2021musiq}
& 66.15 & 6.86
& 62.24 & 9.66
& 71.13 & 7.89
& 68.63 & 13.71 \\

DBCNN~\cite{zhang2018blind}
& 75.52 & 21.64
& 84.09 & \underline{35.90}
& \textbf{85.06} & 30.51
& \textbf{78.50} & 20.12 \\

TReS~\cite{golestaneh2022no}
& \underline{80.67} & \textbf{30.06}
& \underline{86.74} & \textbf{49.88}
& 84.29 & 27.43
& \underline{77.62} & \textbf{28.40} \\

UNIQUE~\cite{zhang2021uncertainty}
& 67.89 & 10.45
& 65.00 & 9.24
& 63.63 & 3.63
& 58.90 & 12.55 \\

LIQE~\cite{zhang2023blind}
& 73.57 & \underline{29.62}
& 77.96 & 32.75
& 82.55 & \underline{34.37}
& 71.81 & \underline{21.66} \\

TOPIQ~\cite{chen2024topiq}
& 75.74 & 11.35
& 75.81 & 22.85
& 80.63 & 24.23
& 71.90 & 14.27 \\
\hline
PrISM-IQA (Ours)
& \textbf{87.85} & 17.80
& \textbf{88.43} & 29.62
& \underline{85.01} & \textbf{35.45}
& 72.42 & 13.66 \\
\bottomrule
\end{tabular}

\begin{tabular}{l|cc|cc|cc|cc}
\toprule
\multirow{2}{*}{Method}
& \multicolumn{2}{c|}{\texttt{wb\_blue\_cast}}
& \multicolumn{2}{c|}{\texttt{wb\_red\_cast}}
& \multicolumn{2}{c|}{\texttt{wb\_yellow\_cast}}
& \multicolumn{2}{c}{mean} \\
\cmidrule(lr){2-3}\cmidrule(lr){4-5}\cmidrule(lr){6-7}\cmidrule(lr){8-9}
& tAUC$\uparrow$ & QWK$\uparrow$
& tAUC$\uparrow$ & QWK$\uparrow$
& tAUC$\uparrow$ & QWK$\uparrow$
& tAUC$\uparrow$ & QWK$\uparrow$ \\
\midrule

MUSIQ~\cite{ke2021musiq}
& 79.40 & 29.23
& 61.92 & 28.32
& 77.35 & 24.83
& 69.55 & 17.21 \\

DBCNN~\cite{zhang2018blind}
& 86.46 & 50.11
& 74.17 & 28.50
& 70.47 & 25.15
& 79.18 & 30.28 \\

TReS~\cite{golestaneh2022no}
& \underline{86.67} & 54.30
& 77.57 & 35.80
& 77.19 & 24.12
& \underline{81.54} & 35.71 \\

UNIQUE~\cite{zhang2021uncertainty}
& 74.96 & 26.41
& 63.47 & 7.84
& 66.98 & 17.13
& 65.83 & 12.47 \\

LIQE~\cite{zhang2023blind}
& 86.19 & \textbf{62.17}
& \underline{84.73} & \underline{44.53}
& \underline{82.31} & \underline{48.40}
& 79.87 & \underline{39.07} \\

TOPIQ~\cite{chen2024topiq}
& 77.89 & 49.72
& 79.21 & 28.70
& 76.14 & 27.64
& 76.76 & 25.54 \\
\hline
PrISM-IQA (Ours)
& \textbf{87.64} & \underline{60.64}
& \textbf{91.81} & \textbf{67.68}
& \textbf{85.12} & \textbf{48.89}
& \textbf{85.47} & \textbf{39.11} \\
\bottomrule
\end{tabular}
\end{table}

\noindent\textbf{Quadratic Weighted Kappa (QWK).}
QWK measures ordinal agreement between \(\hat y_j\) and \(y_j\), with larger penalties for larger severity errors. For issue \(j\), let \(C^{(j)} \in \mathbb{R}^{K_j \times K_j}\) be the confusion matrix:
\[
C_{kk'}^{(j)}
=
\sum_{i \in \mathcal{I}_j}
\mathbb{I}\!\left[y_j^{(i)} = k\right]
\mathbb{I}\!\left[\hat y_j^{(i)} = k'\right],
\qquad
k,k' \in \{1,\dots,K_j\}.
\]
Let
\[
C_{k\bullet}^{(j)} = \sum_{k'=1}^{K_j} C_{kk'}^{(j)},
\qquad
C_{\bullet k'}^{(j)} = \sum_{k=1}^{K_j} C_{kk'}^{(j)}
\]
denote the row and column marginals. The expected count matrix under independent ground-truth and predicted marginals is
\[
E_{kk'}^{(j)}
=
\frac{C_{k\bullet}^{(j)} C_{\bullet k'}^{(j)}}{|\mathcal{I}_j|},
\]
and the quadratic weight between levels \(k\) and \(k'\) is
\[
W_{kk'}^{(j)}
=
\frac{(k-k')^2}{(K_j-1)^2}.
\]
The QWK score is then defined as
\begin{equation}
\mathrm{QWK}_j
=
1
-
\frac{
\sum_{k=1}^{K_j}\sum_{k'=1}^{K_j} W_{kk'}^{(j)} C_{kk'}^{(j)}
}{
\sum_{k=1}^{K_j}\sum_{k'=1}^{K_j} W_{kk'}^{(j)} E_{kk'}^{(j)}
}.
\end{equation}

\noindent\textbf{Aggregation across issues.}
For each reported metric, let \(\mathcal{J} \subseteq \{1,\ldots,N\}\) be the set of issue indices for which that metric is defined and included in the evaluation. We compute each metric per issue and report averages:
\begin{equation}
\mathrm{ACC}
=
\frac{1}{|\mathcal{J}|}
\sum_{j \in \mathcal{J}} \mathrm{ACC}_j,
\qquad
\mathrm{tAUC}
=
\frac{1}{|\mathcal{J}|}
\sum_{j \in \mathcal{J}} \mathrm{tAUC}_j,
\qquad
\mathrm{QWK}
=
\frac{1}{|\mathcal{J}|}
\sum_{j \in \mathcal{J}} \mathrm{QWK}_j.
\end{equation}
When evaluation covers all issues, \(\mathcal{J}=\{1,\ldots,N\}\).

\section{Additional Results on the Expert-Annotated Dataset}
\label{app:aread}
In this section, we provide additional comparisons between PrISM-IQA and adapted NR-IQA baselines on the expert-annotated dataset.

\subsection{Baseline Comparisons}
The main paper reports issue-wise PrISM-IQA results on the expert-annotated dataset in Table~\ref{tab:expert_results}. Here, we compare PrISM-IQA with representative NR-IQA backbones under the same seven-issue ordinal diagnosis setting. For completeness, we adapt existing NR-IQA baselines in the same way as in the main experiments. 
These baselines therefore receive the same issue-level supervision as PrISM-IQA, but they do not use the HEX graph or structured inference layer, isolating the effect of the proposed structured ordinal formulation from the underlying backbone capacity.

Table~\ref{tab:expert_comparisons} shows that PrISM-IQA obtains the best average tAUC. The improvement is most pronounced on white-balance-related labels, including \texttt{wb\_blue\_cast}, \texttt{wb\_red\_cast}, and \texttt{wb\_yellow\_cast}, indicating that the issue-wise ordinal formulation transfers well to expert-labeled chromatic defects. PrISM-IQA is also competitive on tone-related labels such as \texttt{contrast\_high} and \texttt{contrast\_low}.

The QWK gains are less uniform than the tAUC gains, suggesting that exact severity-level agreement remains difficult for some expert-labeled defects. This pattern is consistent with the main observation that tone-related judgments can depend on scene content and subjective tolerance. Overall, the comparisons support the effectiveness of PrISM-IQA against adapted NR-IQA backbones on expert-annotated smartphone photography data.

\begingroup
\footnotesize
\setlength{\tabcolsep}{3pt}
\renewcommand{\arraystretch}{1.08}
\newcolumntype{L}[1]{>{\RaggedRight\arraybackslash}p{#1}}

\begin{longtable} {@{}cL{0.10\linewidth}L{0.24\linewidth}L{0.24\linewidth}L{0.32\linewidth}@{}}
\caption{Issue-to-module rules for PrISM-IQA-guided OpenISP tuning. Each issue is assigned a primary OpenISP tuning module. Auxiliary modules may be activated for stronger correction or joint adjustment when the predicted issue is severe or coupled with related perceptual failures.}
\label{tab:openisp_issue_module_map}\\
\toprule
ID & Scope & Issue & Primary module & Auxiliary modules \\
\midrule
\endfirsthead

\toprule
ID & Scope & Issue & Primary module & Auxiliary modules \\
\midrule
\endhead

\bottomrule
\endfoot

1 & \code{building} & \code{brightness_high} & \code{contrast_brightness_control} & \code{gamma_correction} \\
2 & \code{building} & \code{brightness_low} & \code{contrast_brightness_control} & \code{gamma_correction} \\
3 & \code{building} & \code{clarity_low} & \code{cfa_interpolation} & \code{noise_filter_for_luma}, \code{edge_enhancement} \\
4 & \code{building} & \code{contrast_high} & \code{contrast_brightness_control} & \code{gamma_correction} \\
5 & \code{building} & \code{contrast_low} & \code{contrast_brightness_control} & \code{gamma_correction} \\

6 & \code{face} & \code{brightness_high} & \code{contrast_brightness_control} & \code{black_level_compensation}, \code{gamma_correction} \\
7 & \code{face} & \code{brightness_low} & \code{contrast_brightness_control} & \code{black_level_compensation}, \code{gamma_correction} \\
8 & \code{face} & \code{clarity_low} & \code{cfa_interpolation} & \code{anti_aliasing_noise_filter}, \code{noise_filter_for_luma}, \code{edge_enhancement} \\
9 & \code{face} & \code{contrast_high} & \code{contrast_brightness_control} & \code{gamma_correction} \\
10 & \code{face} & \code{contrast_low} & \code{contrast_brightness_control} & \code{gamma_correction} \\
11 & \code{face} & \code{hair_detail_loss} & \code{noise_filter_for_luma} & \code{anti_aliasing_noise_filter}, \code{cfa_interpolation}, \code{edge_enhancement} \\
12 & \code{face} & \code{highlight_clipped} & \code{contrast_brightness_control} & \code{awb_gain_control}, \code{gamma_correction}, \code{color_correction_matrix} \\
13 & \code{face} & \code{highlight_muted} & \code{gamma_correction} & \code{contrast_brightness_control} \\
14 & \code{face} & \code{lip_color_cast} & \code{color_correction_matrix} & \code{awb_gain_control}, \code{hue_saturation_control} \\
15 & \code{face} & \code{noise} & \code{noise_filter_for_luma} & \code{anti_aliasing_noise_filter}, \code{noise_filter_for_chroma} \\
16 & \code{face} & \code{sharpening_artifact} & \code{edge_enhancement} & \code{cfa_interpolation}, \code{false_color_suppression} \\
17 & \code{face} & \code{skin_blue_cast} & \code{awb_gain_control} & \code{color_correction_matrix}, \code{hue_saturation_control} \\
18 & \code{face} & \code{skin_green_cast} & \code{awb_gain_control} & \code{color_correction_matrix}, \code{hue_saturation_control} \\
19 & \code{face} & \code{skin_red_cast} & \code{awb_gain_control} & \code{color_correction_matrix}, \code{hue_saturation_control} \\
20 & \code{face} & \code{skin_yellow_cast} & \code{awb_gain_control} & \code{color_correction_matrix}, \code{hue_saturation_control} \\
21 & \code{face} & \code{texture_artifact} & \code{edge_enhancement} & \code{cfa_interpolation}, \code{noise_filter_for_luma} \\

22 & \code{greenery} & \code{clarity_low} & \code{cfa_interpolation} & \code{noise_filter_for_luma}, \code{edge_enhancement} \\
23 & \code{greenery} & \code{color_cast} & \code{awb_gain_control} & \code{color_correction_matrix}, \code{hue_saturation_control} \\
24 & \code{greenery} & \code{contrast_high} & \code{contrast_brightness_control} & \code{gamma_correction} \\
25 & \code{greenery} & \code{contrast_low} & \code{contrast_brightness_control} & \code{gamma_correction} \\
26 & \code{greenery} & \code{saturation_high} & \code{hue_saturation_control} & \code{color_correction_matrix} \\
27 & \code{greenery} & \code{saturation_low} & \code{hue_saturation_control} & \code{color_correction_matrix} \\

28 & \code{sky} & \code{color_cast} & \code{awb_gain_control} & \code{lens_shading_correction}, \code{color_correction_matrix}, \code{hue_saturation_control} \\
29 & \code{sky} & \code{compositing_artifact} & \code{cfa_interpolation} & \code{false_color_suppression} \\
30 & \code{sky} & \code{contrast_high} & \code{contrast_brightness_control} & \code{gamma_correction} \\
31 & \code{sky} & \code{contrast_low} & \code{contrast_brightness_control} & \code{gamma_correction} \\
32 & \code{sky} & \code{highlight_clipped} & \code{gamma_correction} & \code{contrast_brightness_control} \\
33 & \code{sky} & \code{highlight_muted} & \code{gamma_correction} & \code{contrast_brightness_control} \\
34 & \code{sky} & \code{saturation_high} & \code{hue_saturation_control} & \code{color_correction_matrix} \\
35 & \code{sky} & \code{saturation_low} & \code{hue_saturation_control} & \code{color_correction_matrix} \\

36 & \code{dominant_color} & \code{color_cast} & \code{awb_gain_control} & \code{color_correction_matrix}, \code{hue_saturation_control} \\
37 & \code{dominant_color} & \code{saturation_high} & \code{hue_saturation_control} & \code{color_correction_matrix} \\
38 & \code{dominant_color} & \code{saturation_low} & \code{hue_saturation_control} & \code{color_correction_matrix} \\

39 & \code{global} & \code{brightness_high} & \code{contrast_brightness_control} & \code{black_level_compensation}, \code{lens_shading_correction}, \code{gamma_correction} \\
40 & \code{global} & \code{brightness_low} & \code{contrast_brightness_control} & \code{black_level_compensation}, \code{lens_shading_correction}, \code{gamma_correction} \\
41 & \code{global} & \code{clarity_low} & \code{cfa_interpolation} & \code{anti_aliasing_noise_filter}, \code{noise_filter_for_luma}, \code{edge_enhancement} \\
42 & \code{global} & \code{contrast_high} & \code{contrast_brightness_control} & \code{gamma_correction} \\
43 & \code{global} & \code{contrast_low} & \code{contrast_brightness_control} & \code{gamma_correction} \\
44 & \code{global} & \code{highlight_clipped} & \code{contrast_brightness_control} & \code{awb_gain_control}, \code{gamma_correction}, \code{color_correction_matrix} \\
45 & \code{global} & \code{highlight_muted} & \code{gamma_correction} & \code{contrast_brightness_control} \\
46 & \code{global} & \code{noise} & \code{noise_filter_for_luma} & \code{dead_pixel_correction}, \code{anti_aliasing_noise_filter}, \code{noise_filter_for_chroma} \\
47 & \code{global} & \code{sharpness_high} & \code{edge_enhancement} & \code{cfa_interpolation} \\
48 & \code{global} & \code{wb_blue_cast} & \code{awb_gain_control} & \code{lens_shading_correction}, \code{color_correction_matrix}, \code{hue_saturation_control} \\
49 & \code{global} & \code{wb_green_cast} & \code{awb_gain_control} & \code{lens_shading_correction}, \code{color_correction_matrix}, \code{hue_saturation_control} \\
50 & \code{global} & \code{wb_red_cast} & \code{awb_gain_control} & \code{lens_shading_correction}, \code{color_correction_matrix}, \code{hue_saturation_control} \\
51 & \code{global} & \code{wb_yellow_cast} & \code{awb_gain_control} & \code{lens_shading_correction}, \code{color_correction_matrix}, \code{hue_saturation_control} \\

\end{longtable}
\endgroup

\end{document}